\documentclass{article}

\usepackage{microtype}
\usepackage{graphicx}
\usepackage{subfigure}
\usepackage{booktabs} 
\usepackage{multirow}
\usepackage{soul} 

\usepackage{hyperref}

\usepackage[accepted]{icml2024}

\usepackage{amsmath}
\usepackage{amssymb}
\usepackage{mathtools}
\usepackage{amsthm}

\usepackage[capitalize,noabbrev]{cleveref}

\theoremstyle{plain}
\newtheorem{theorem}{Theorem}[section]

\theoremstyle{definition}
\newtheorem{definition}[theorem]{Definition}

\theoremstyle{remark}

\usepackage[textsize=tiny]{todonotes}

\icmltitlerunning{To Each (Textual Sequence) Its Own: Improving Memorized-Data Unlearning in Large Language Models}

\begin{document}

\twocolumn[
\icmltitle{To Each (Textual Sequence) Its Own: \\ 
Improving Memorized-Data Unlearning in Large Language Models}



\icmlsetsymbol{equal}{*}

\begin{icmlauthorlist}
\icmlauthor{George-Octavian Bărbulescu}{yyy}
\icmlauthor{Peter Triantafillou}{yyy}
\end{icmlauthorlist}

\icmlaffiliation{yyy}{Department of Computer Science, University of Warwick, United Kingdom}

\icmlcorrespondingauthor{George-Octavian Bărbulescu}{george-octavian.barbulescu@warwick.ac.uk}

\icmlkeywords{Machine Learning, ICML}

\vskip 0.3in
]



\printAffiliationsAndNotice{}  

\begin{abstract}

LLMs have been found to memorize training textual sequences and regurgitate verbatim said sequences during text generation time. This fact is known to be the cause of privacy and related (e.g., copyright) problems.
Unlearning in LLMs then takes the form of devising new algorithms that will properly deal with these side-effects of memorized data, while not hurting the model's utility. We offer a fresh perspective towards this goal, namely, that each textual sequence to be forgotten should be treated differently when being unlearned based on its degree of memorization within the LLM. We contribute a new metric for measuring unlearning quality, an adversarial attack showing that SOTA algorithms lacking this perspective fail for privacy, and two new unlearning methods based on Gradient Ascent and Task Arithmetic, respectively. A comprehensive performance evaluation across an extensive suite of NLP
tasks then mapped the solution space, identifying the best solutions under different scales in model capacities and forget set sizes and quantified the gains of the new approaches.
\end{abstract}

\section{Introduction}
\label{intro}

LLMs are models typically built over transformer architectures and trained on a vast corpus of data (with up to trillions of tokens) \cite{radford2019language,chowdhery2023palm, gao2020pile, fedus2022switch, kassem2023preserving}. Massive training data and a (lengthy) training process allow  LLMs to establish factual associations and memorize language semantics and grammar \cite{zhang2021counterfactual}. 

Unfortunately, previous research has shown that LLMs memorized training examples which causes them to emit information verbatim as it exists in the training corpus when prompts are carefully designed. This, in turn, violates privacy regardless of any sophisticated adversarial attacks. 
These problems are juxtaposed against laws and regulations protecting the rights of individuals to be forgotten (RTBF)
\cite{mantelero2013eu, graves2021amnesiac}, that is, to have their information removed from the LLM. 
A straightforward approach to achieve this would be to retrain the LLM from `scratch'.
Alas, this is infeasible in the context of LLMs, even with small/medium-sized models (millions of parameters). The problem is further exacerbated by the fact that the tendency of LLMs to regurgitate training data increases proportionally with the number of model parameters \cite{carlini2022quantifying}. 
Hence, the need to research efficient removal mechanisms of memorized data from pretrained LLMs, without requiring training the LLMs from scratch.

As mentioned, LLMs have been shown to (i) memorize and (ii) emit memorized training data at generation time, which causes privacy and copyright problems. 
As an example, GPT-3 \cite{brown2020language} has been shown to generate PII information verbatim, raising significant concerns given its wide commercial usage \cite{heikkila2022does}. 
Interestingly, memorized training data points can also decrease the quality of the model \cite{zhang2021counterfactual}, by linking tokens which do not have a general meaning or a justifiable factual association, but only belong to an entity or an individual.
But, given that memorization is the root cause of these problems, how can one intuitively define 'memorization'? For LLMs, typically, if there exists a sufficient overlap between a training data point and a generated text response, 
then that data point is considered to have been memorized \cite{tirumala2022memorization, carlini2021extracting, carlini2022quantifying}. 

The focus of this research is to enrich LLMs with the ability to unlearn memorized textual sequences (a.k.a. training data points, examples, or samples), while satisfying three constraints: First, do so 
while maintaining the model's utility for downstream applications unrelated to the forgotten information. 
Second, do so without having access to the rest of the pre-training dataset, as this may be either intractable or too large/costly to fine-tune over.
Third, do so while avoiding the aforementioned problems with respect to privacy, copyrights, etc.
As a real-world example, consider the need to remove the effect of one or more textual sequences from the LLM for copyright reasons \cite{graves2021amnesiac, villaronga2018humans}. One approach to address this problem would be to place constraints on the original data (e.g. to obfuscate it) at inference time \cite{majmudar2022differentially, lu2020neurologic}. However, such a solution would not protect from adversarial attacks to expose PII; to deal with such attacks, one must also erase the contribution of the targetted textual sequences from the model's parameters. To protect against adversaries with white-box access to the model's parameters, the community has developed algorithms that ensure differential privacy (DP) \cite{abadi2016deep, yu2021differentially, li2021large}. Unfortunately, it has been discovered that training LLMs with DP can be very expensive \cite{anil2021large}. 
Also, 
given that the memorized information that needs to be removed 
is not known during training, it is impossible to draw a clear privacy boundary \cite{brown2022does}, and training with DP on the entire training dataset introduces noise which hurts model utility \cite{anil2021large, nguyen2022survey}.

We focus on the problem of deleting memorized textual sequences from the models' weights, as we deem it is a fundamental first step in unlearning in generative language models. 
We claim that there exists no satisfactory solution to this fundamental first step in unlearning in LLMs, as it depends, on the one hand, on the number of samples (a.k.a. data points, text sequences) to be forgotten, and, on the other hand, on the capacity (number of parameters) of the model. Existing state-of-the-art solutions include self-supervised learning techniques \cite{jang2022knowledge, liu2023forgetting, chen2023unlearn} and reinforcement learning methods \cite{kassem2023preserving, lu2022quark}. 
To quantify the success of unlearning, previous methods holistically characterize the memorization of the whole {\it forget set}, that is, the set of textual sequences that are to be `unlearned'. 
Hence, the forget set is given a "memorization score" which is the average of the memorization scores of each textual sequence \cite{jang2022knowledge}. If the average score is within an expected threshold, then the unlearning algorithm is deemed successful. 
Such expected thresholds are defined as the level of memorization one would expect to see for arbitrary text sequences unseen during training.
We hypothesized and show with this work that looking at average memorization scores is not sufficient to conclude that forget-set examples are no longer extractable at generation time. This is in principle akin to the finding by Carlini et al that average values are misleading when defending against Membership Inference Attacks (MIAs) \cite{carlini2022membership}.
In our problem setting, intuitively, the distribution of memorization scores may be non-uniform (and indeed long-tailed). As such, it is possible that a subpopulation of textual sequences in the forget set is still memorized, while another subpopulation may have been completely forgotten. This leads 
to important privacy loopholes: (i) highly memorized data points can be easily extracted; and (ii) even data points that are unlearned ``too well" can still be extractable with a minimum-effort white-box access to the LLM. Alas, current state-of-the-art solutions are susceptible to these phenomena.

In light of the above, we propose to quantify the success of unlearning by looking at sequence-level memorization. We quantify unlearning success by counting the number of outliers with respect to a (desired) memorization threshold. For example, if the average memorization score across a large population of unseen (during training) textual sequences is, say, 30\%, then the subpopulation of the forget set that scores above 60\% after unlearning is still extractable. 
Furthermore, we propose two unlearning algorithms designed to meet this new criterion. 
The key aim behind them lies in providing fine-grained (per textual sequence) memorization control.

{\bf Contributions}:
\vspace{-0.3cm}
\begin{itemize}
\setlength{\itemsep}{1pt}
  \setlength{\parskip}{0pt}
  \setlength{\parsep}{0pt}
    \item We introduce a new metric to quantify the success of forgetting textual sequences in LLMs, which focuses on example-level memorization as opposed to aggregate memorization across forget-set examples. 
    \item We devise a new Membership Inference Attack (MIA) for unlearning memorized data in LLMs, which shows evidence that existing SOTA unlearning algorithms, that focus on the aggregate memorization in the forget-set examples, are prone to privacy violations as a result of the presence of under- and over-memorized subpopulations of data points after unlearning. 
    \item We introduce a novel metric to quantify model indistinguishability for unlearning algorithms in LLMs. Intuitively, indistinguishability is the closeness in the behaviour of two models: (i) the unlearned model and (ii) models retrained from scratch without the textual samples to be removed. To map behaviour, we inspect the model’s loss, which is a principled approach in SOTA methods from privacy literature.
    \item We introduce two (fairly simple) algorithms that provide the desired fine-grained, per-example unlearning.  
    \item The two new algorithms piggyback upon known SOTA algorithms for unlearning in LLMs, and for detoxifying LLMs. Interestingly, we show that a solution for detoxifying LLMs can become a SOTA solution for unlearning in LLMs in specific circumstances.
    \item We study the solution space across model and forge-set sizes using nine NLP classification  and four text generation tasks. We identify the best approach in different circumstances. We compare the new algorithms against existing SOTA unlearning solutions in LLMs, showing their superiority for both model utility and privacy metrics. So, in this case,``simple does it"! 
\vspace{-0.3cm}
\end{itemize}

Put together, we hope the above constitute actionable and significant steps towards understanding and solving memorized-data unlearning in LLMs.

\section{Background}

\subsection{Memorization in Large Language Models}

Memorization in LLMs has been studied extensively as it comes with significant privacy and legal concerns 
\cite{carlini2021extracting, carlini2022quantifying, jagielski2022measuring, tirumala2022memorization}. 
Pretrained LLMs have a tendency to regurgitate verbatim some of the training data, and their memorization is proportional to their number of parameters \cite{carlini2021extracting, lee2021deduplicating}. For instance, GPT-J which features 6 billion parameters has been shown to have memorized at least 1\% \cite{carlini2022quantifying, mesh-transformer-jax, black2021gpt} of Pile \cite{gao2020pile}, the dataset used during its training. This memorization, albeit unintended, is crucial to model natural language, and for generalisation \cite{zhang2021counterfactual, feldman2020does}.
A range of metrics have been proposed to compute the level of memorization across data points in causal language modelling \cite{tirumala2022memorization,carlini2022quantifying,jagielski2022measuring}, based on the likelihood of emitting textual sequences verbatim. 

\subsection{Machine Unlearning for Memorized Data}
Machine unlearning has been receiving increasing attention due primarily to ensuring RTBF, but also for other applications \cite{kurmanji2023towards} - albeit unlearning research for image classification has matured more than unlearning in LLMs (see for instance the NeurIPS 2023 unlearning competition \cite{unlearn_competition}). 
Unlearning memorized data from LLMs is the field of study concerned with efficient methods for removing the contributions of training examples from the model \cite{maini2024tofu, neel2023privacy, zhang2024comprehensive}. Exact unlearning retrains LLMs from scratch by first processing the dataset (i.e., removing the training data points requested to be removed). However, this is a costly process, especially when it comes to retraining very large transformers, the current state-of-the-art architecture in NLP \cite{brown2020language, abadi2016deep}. To mitigate the costs associated with retraining LLMs for each forget request, approximate unlearning methods have been proposed \cite{pawelczyk2023context, chen2023unlearn, meng2022locating, jang2022knowledge, kassem2023preserving}
which remove the contribution of the target data points from the pretrained model. 

\section{Related Work}
\label{related}

Related works can be taxonomized based on whether they: 
(i)  require access to training datasets; 
(ii) target language generation;
(iii) alter model weights or simply alter LLM output; and
(iv)  target unlearning memorized data or `higher-level' entities such as entity or factual associations. 

We focus on: 
\textbf{(i) algorithms which avoid accessing the pretraining data}. These are solutions which assume that retain data (training data minus forget set) is too large/costly to be acted upon,
in contrast to unlearning research leveraging training datasets \cite{chen2023unlearn, liu2023forgetting};
\textbf{(ii) on language generation}, unlike unlearning methods designed for text classification (e.g., via \textit{label flipping} \cite{pawelczyk2023context});
\textbf{(iii) on methods that alter model weights}. Differential Privacy Decoding (DPD) methods (e.g., \cite{majmudar2022differentially}) are inference-time techniques. They remain relevant to this work since they could be used to solve, for instance, copyright issues, oblivious to the fact that they do not alter model weights. Nonetheless, we will compare against DPD to showcase its privacy problems;
\textbf{(iv) on unlearning memorized training text}, instead of removing factual associations or higher-level semantics such as bias and toxicity \cite{meng2022locating, wang2023knowledge, lu2022quark, kassem2023preserving}. For instance, \cite{kassem2023preserving} derive a paraphrasing policy by repeatedly sampling answers from the original LLM with respect to parts of the original data points (used as prompts), picking the sample least similar to the original. 
Akin to detoxifying or removing biases from LLMs, the method mitigates \textit{approximate memorization}, i.e., higher-level similarity to the original training text \cite{ippolito2022preventing}, instead of \textit{exact memorization} \cite{tirumala2022memorization}, namely, the likelihood of extracting or emitting a training textual sequence verbatim, which is a closer concern to data copyright and adversarial attacks.

The current SOTA approach for our task at hand (also regarded as SOTA in  \cite{liu2023forgetting, kassem2023preserving, pawelczyk2023context} 
is the work of \cite{jang2022knowledge}, who uses gradient ascent (GA) oblivious to the pretraining dataset to minimize exact memorization over a set of textual sequences. It shows that simply maximizing the loss with respect to the textual sequences to be forgotten produces a competitive trade-off between privacy and model utility. We will compare against this SOTA algorithm quantifying further improvements. Finally, Task Arithmetic (TA) \cite{ilharco2022editing} is a recent learning paradigm that builds on the intuition that learning a new task can be seen as a vector of weights in neural networks. Task arithmetic has shown impressive results for removing biases and detoxifying LLMs. Although these are very similar tasks to unlearning, 
using TA in LLMs for unlearning has never been investigated before. We will show this is unfortunate.

\section{The New Approach}

\subsection{Quantifying Memorization in LLMs}

To quantify the dynamics of memorization during unlearning we adopt a variation of \textit{label memorization} \cite{zhang2021understanding, pondenkandath2018leveraging} produced by \cite{tirumala2022memorization} to study the properties of neural LMs during training, referred to as \textit{exact memorization}. 
The motivation for using \textit{exact memorization} to characterise  LLMs' knowledge over a textual sequence is twofold: 1) it is prominent in previous work that explores copyright issues in LLMs \cite{jang2022knowledge, tirumala2022memorization}, and 2) it works well in practical settings as it is inexpensive to compute. 

\begin{definition}
Let $x=(x_1,x_2,..,x_T)$ be a textual sequence of $T$ tokens with $x\in X$, where $X$ is the set of all possible textual sequences over the LMs' vocabulary. Let $f_\theta$ be a neural LM with parameters $\theta$. Then \textit{exact memorization} is a function $g: X \rightarrow \mathbb{R}^+$ where:
\begin{equation}
\vspace{-0.4cm}
\label{memeq}
    g(x) = \frac{\sum_{i=1}^{T-1} \mathbf{1}\{ f_\theta(\cdot | x_{<i}) = x_i \} } {T-1}
\end{equation}
\end{definition}

\textit{Exact memorization} is used in prior research to define/compute the success of forgetting in LLMs.  

\begin{definition}
 Let $D_f \subset X$ be the set of textual sequences to be forgotten 
 and $D_t \subset X, D_t \cap D_f = \emptyset$, be a test set of textual sequences (i.e. text not used during the training of the model), then a \textit{forgetting} algorithm is successful from a memorization standpoint if the following inequality holds:
\begin{equation}
\frac{1}{|D_f|}\sum_{x_f \in D_f}g(x_f) \leq \frac{1}{|D_t|} \sum_{x_t \in D_t}g(x_t)
\label{ineq}
\end{equation}
\end{definition}

\begin{figure}[t]
\vspace{-0.4cm}
\vskip 0.2in
\begin{center}
\centerline{\includegraphics[scale=0.35]{ 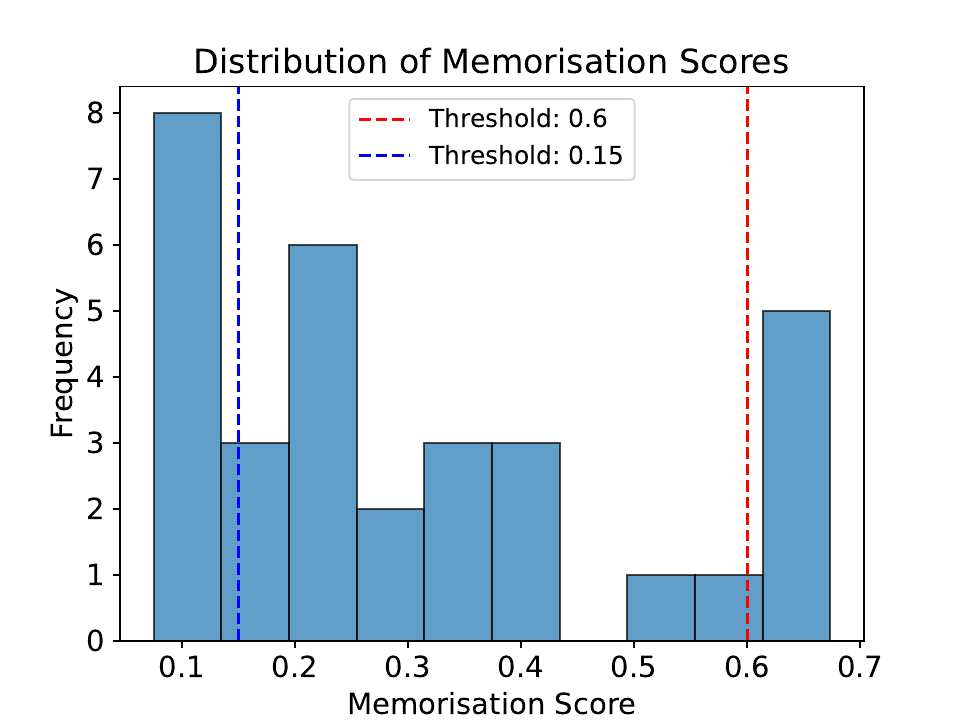}}
\vspace{-0.4cm}
\caption{The distribution of \textit{exact memorization} scores across the forget set $D_f$ when the average memorization threshold is $\leq 33\%$ (as exhibited by GPT-Neo with 1.3B parameters over unseen/arbitrary textual sequences \cite{jang2022knowledge}). While \cref{ineq} holds, 5 (8) samples from $D_f$ have an exact memorization score of over $60\%$ (under $10\%)$. 
}
\label{memo_hist}
\end{center}
\vskip -0.2in
\vspace{-0.6cm}
\end{figure}

The intuition behind the unlearning success threshold is that the population of textual sequences in the forget set $D_f$ should be characterized by a memorization score lower or similar to that of the population of samples unseen during the LLMs training (see \cref{sample-table} for forgetting thresholds across the GPT-Neo family \cite{black2021gpt}). 

However, we intuit that an average memorization score over the whole forget set will in general not provide enough information needed to deem an unlearning algorithm's execution as successful and is thus inappropriate. This is in principle
akin to Carlini et al's argument that average values
are inappropriate to judge successful defence against Membership Inference Attacks (MIAs) \cite{carlini2022membership}. 
We showcase this with an experiment we conducted where the forget set $D_f$ is composed of 32 textual sequences of $T=200$ tokens, each picked arbitrarily from the Pile corpora, known to be easily extractable from the GPT-Neo family of models. Then, we apply gradient ascent (i.e., we maximise $\mathcal{L}_\theta(x) = - \sum_{t=1} log f_\theta(x_t | x_{<t})$, where $x \in D_f$) to unlearn the textual sequences. The distribution of the memorization scores in $D_f$ after unlearning is shown in \cref{memo_hist}.

The figure depicts a clear non-uniform and highly variable distribution of the memorization scores across examples in $D_f$. 
Note that the algorithm has met the empirical threshold for unlearning success. 
Nonetheless, as discussed earlier (and as we will show later) such highly variable memorization scores of examples in $D_f$ constitute loopholes that can lead to copyright liabilities and/or can be exploited to reveal private information.  
While some variation in memorization is to be expected across $D_f$, as some data points contain more common or generalizable n-grams, large variations in memorization scores are inherently problematic. 
This led our research to the question of whether there exists a class of unlearning algorithms that can further reduce the likelihood of \textit{memorization-score outliers} (that is, training data points with very low or very high memorization scores during unlearning) which will ensure lower susceptibility to privacy attacks and copyright violations and without impacting model utility. 
For this we propose to use the following metric which simply counts such outliers after unlearning (i.e., when \cref{ineq} becomes true). 
\begin{equation}
\vspace{-0.3cm}
\label{counteq}
c(D_f) = \sum_{x \in D_f} \mathbf{1}\{g(x) \in ([0, lb] \cup [ub, 1])\}
\end{equation}
\cref{counteq} counts the number of points with memorization scores after unlearning, $m$, where 
$m \ < \ lb$ or $ub \ < \ m$.  
The lower and upper bound thresholds $lb$ and $up$ are found empirically and depend on the training data set.
In our case, for example, we have chosen empirically $lb = 15\%$ and $ub = 60\%$.  
The motivation for counting data points with very low memorization scores is that these become extractable via adversarial attacks (as first shown using the Streisand effect in \cite{golatkar2020eternal}), and we hypothesize these also impact model utility (e.g. by unlearning useful, recurrent n-grams). We have tested this hypothesis by performing a membership inference attack on the unlearned textual sequences (see \cref{streissand}) and by measuring model indistinguishability on these (see \cref{loss magnitude}).

\subsection{Selective Unlearning}

Building on our insight that reducing memorization for privacy is a data-point-level problem and that unlearning success cannot be appropriately measured with an aggregate over the forget set examples (which is the approach commonly taken in computer vision unlearning \cite{kurmanji2023towards} and in previous unlearning research in  LLMs \cite{jang2022knowledge, kassem2023preserving}), we propose two alternative unlearning algorithms to provide fine-grained exact memorization control. 

\subsubsection{Selective Gradient Ascent}
En route to devising our first algorithm, we visit gradient ascent (GA). 
GA is an ever-present method for unlearning in various settings. It is found, for instance, in unlearning in image classification literature as "NegGrad" in \cite{kurmanji2023towards} and in forgetting text in LLMs \cite{jang2022knowledge}. Gradient ascent works by simply maximizing the loss function for examples in the forget set, instead of minimising it, where the loss function is dependent on the task. In causal language modelling, unlearning minimizes: 
\begin{equation}
\vspace{-0.3cm}
\label{ga_eq}
    \mathcal{L}_\theta(x) = \sum_{t=1}^{T-1} log f_\theta(x_t | x_{<t}), \text{where }x \in D_f
\end{equation}
Note \cref{ga_eq} is the same as maximising the original log-likelihood function used in causal language modelling ($\mathcal{L}_\theta(x)=-\sum_{t=1}^{T-1} log f_\theta(x_t | x_{<t})$, where $x = (x_1, x_2, ..., x_T)$ is a sequence of tokens). If the number of examples in the forget set is larger than one, gradient ascent becomes oblivious to the individual memorization scores and stops when the average memorization across the forget set meets the empirical threshold for unlearning success. 

To control the level of memorization we introduce Selective Gradient Ascent (SGA), which performs a two-step optimisation routine. Let $\gamma$ be an upper limit for the acceptable exact memorization score of any given sample from the forget set after the process of unlearning, and let $D_\gamma$ be the of data points above the limit: $D_\gamma = \{x | g(x) > \gamma, x \in D_f\}$, where $g(x)$ computes the memorization score at the example level as per \cref{memeq}. The unlearning algorithm begins by performing gradient ascent with respect to $D_\gamma$. After each epoch, $D_\gamma$ is recomputed based on the entire forget set $D_f$. If there are no more elements above the unacceptable memorization threshold, $D_\gamma =\emptyset$, and inequality \cref{ineq} is false, we continue unlearning with gradient ascent with respect to the top-k memorized samples from the forget set,  $D_{tk} = \text{arg\,top}_k\{ (x,g(x)) | x \in D_f\}$, recomputed after each epoch until the average memorization across the population of textual sequences reaches the desired level.

\textbf{Discussion:} SGA aims to provide fine-grained unlearning. 
The same cannot be achieved via sequential gradient ascent, i.e., unlearning each sequence one by one, as the data points in the forget set may share n-grams and their memorization scores may be linked. In this case, the order in which to unlearn data points is crucial and requires considering all possible orderings and suffering the associated combinatorial explosion in time. In this research, we merely touch the surface of our proposed paradigm, \textit{selective unlearning}, which can be adapted to any gradient-based method. 
 
\subsubsection{Task Arithmetic for Unlearning}
\begin{table}[t]
\caption{The forgetting thresholds for GPT-Neo models on the Pile validation corpora, from \cite{jang2022knowledge}. Average memorization across 10,000 unseen textual sequences.}
\vspace{-0.5cm}
\label{sample-table}
\vskip 0.15in
\begin{center}
\begin{small}
\begin{sc}
\begin{tabular}{lcccr}
\toprule
Number of Parameters & Memorization Threshold \\
\midrule
GPT-Neo (125M) & 29.94\% \\
GPT-Neo (1.3B) & 33.27\% \\
GPT-Neo (2.7B) & 34.02\% \\
\bottomrule
\end{tabular}
\end{sc}
\end{small}
\end{center}
\vskip -0.1in
\vspace{-0.5cm}
\end{table}

\begin{table*}[t]
\caption{Performance overview across the suite of unlearning algorithms. Unlearning 16 samples. RUNTIME is shown in minutes. At least one of the new algorithms we present (TA, TAU, and/or SGA) outperform GA for F1, CE, and ACC. For EXTRACTABLE, SGA is significantly superior to all, followed by TA/TAU, while GA has poor performance.}
\vspace{-0.4cm}
\label{metrics_table}
\vskip 0.15in
\begin{center}
\begin{small}
\begin{sc}
\begin{tabular}{lccccccc}
\toprule
\multirow{2}{*}{Model} & \multirow{2}{*}{Algo} & \multirow{2}{*}{F1 $\uparrow$} & \multirow{2}{*}{CE $\downarrow$} & \multirow{2}{*}{ACC $\uparrow$} & \multirow{2}{*}{EXTRACTABLE $\downarrow$} & \multirow{2}{*}{RUNTIME $\downarrow$} & \multirow{2}{*}{$\Delta$MEM $\downarrow$} \\
& & & & & & & \\
\midrule
GPT-Neo(125M) 
       & DEFAULT  & 10.27 & 3.82 & 44.44 & 13.8 & n/a & 50.14 \\
& DPD & 6.23  & n/a & n/a & 8.8 & n/a & 15.44 \\

& GA  & 2.28 & 6.18 & 40.34 & 5.0 & 7.22 & 3.34 \\
         & SGA & 1.86 & 5.63 & 41.67 & 1.8 & 8.18 & 2.78 \\
         & TA  & 7.18 & 4.82 & 39.51 & 2.2 & 7.91 & 0.55 \\
         & TAU & 7.37 & 4.44 & 43.68 & 3.4 & 50.74 & 6.18 \\

\midrule
GPT-Neo(1.3B) 
       & DEFAULT  & 12.86 & 3.29 & 51.39 & 16.0 & n/a & 60.34 \\
       & DPD & 6.52  & n/a & n/a & 2.4 & n/a & 10.95 \\

& GA  & 11.34 & 3.52 & 49.37 & 4.6 & 16.23 & 4.06 \\
         & SGA & 12.63 & 3.37 & 49.30  & 1.0 & 20.17 & 1.86 \\
         & TA  & 9.60  & 3.69 & 48.75 & 2.0 & 14.80 & 1.80 \\
         & TAU & 12.52 & 3.39 & 49.24 & 4.0 & 29.96 & 0.81 \\
         & GA+Clip & 10.22 & 3.64 & 50.42 & 3.6 & 18.19 & 1.76 \\
\midrule
GPT-Neo(2.7B) 
       & DEFAULT  & 11.23 & 3.09 & 56.94 & 16.0 & n/a & 60.55 \\
       & DPD & 6.27  & n/a & n/a & 1.0 & n/a & 3.73 \\
        & GA  & 10.99 & 3.27 & 56.25 & 7.0 & 22.14 & 6.22 \\
         & SGA & 10.94 & 3.32 & 57.22 & 0.8 & 29.96 & 4.28 \\
         & TA  & 10.03 & 3.81 & 49.58 & 1.6 & 19.95 & 2.49 \\
         & TAU & 11.34 & 3.27 & 55.14 & 4.0 & 36.54 & 1.52 \\

\bottomrule
\end{tabular}
\end{sc}
\end{small}
\end{center}
\vskip -0.1in
\end{table*}
A contrasting approach to unlearning information is to identify in the model's parameter space the weights related to the forget set. This is a technique that has been used for unlearning in image classification settings \cite{golatkar2020eternal, thudi2022unrolling, NEURIPS2022_5771d9f2, chundawat2023can, kurmanji2023towards}, as well as in LLMs.
For LLMs, one technique to achieve this is to observe which parameters change when fine-tuning the model on the forget set. A technique that adopts this paradigm is task arithmetic \cite{ilharco2022editing}. 
\begin{definition}
    Let $\theta \in \mathbb{R}^d$ be the parameters of a neural language model, where $d$ is the number of parameters. Let $\mathcal{A}$ be a deep learning algorithm such as fine-tuning with gradient descent. Then $\theta_{ft} = \mathcal{A}(\theta, D_f)$ are the model parameters after fine-tuning on the forget set. To forget this information, \textit{task arithmetic} is expressed as:
\begin{equation}
\label{ta_forget}
\theta_u  = \theta - (\theta_{ft} - \theta)
\vspace{-0.3cm}
\end{equation}
\end{definition}

The intuition behind task arithmetic is to identify the direction in which the model's parameters move when learning a new task. This is analogous to identifying the important weights for the task, which form a vector $\delta_{ft} =(\theta_{ft} - \theta)$. While removing this task vector from the original model as per \cref{ta_forget} has shown promise in detoxifying neural LMs \cite{ilharco2022editing}, applying it for our unlearning task is not trivial. The forget set $D_f$ is generally comprised of data points that are highly memorized, otherwise, there would be no need for unlearning in the first place. Therefore, it is crucial to reflect on whether the drift $\delta_{ft}$ is a good approximation of the learning movement that occurred when $D_f$ was initially memorized.

We hypothesize 
the `drift' when the level of memorization across the forget set is already very high (with respect to \cref{memeq}) cannot be used for unlearning purposes. The intuition is that the data is already learned well, and further fine-tuning on the forget set may lead to small changes in the parameter space. Furthermore, even if the average memorization score is not high, there could be outliers (see Figure \ref{memo_hist}), and these highly-memorized members become impossible to unlearn for the same reason. 

Therefore, we first run a couple of epochs using SGA with a twofold objective in mind: 1) to standardize the distribution of memorization scores across the forget set, and 2) to decrease the average memorization. 
\begin{definition}
    Let $\theta_{sga} \in \mathbb{R}^d$ be the weights of the target model after applying SGA for a number of epochs $k$ and let $\lambda$ be a hyperparameter to adjust the size of the vector, then Task Arithmetic for Unlearning (TAU) is defined as:
    \vspace{-0.2cm}
    \begin{equation}
        \theta_{u} = \theta_{sga} - \lambda(\mathcal{A}(\theta_{sga}, D_f) - \theta_{sga})
    \vspace{-0.3cm} 
    \end{equation}
 
\end{definition}

\section{Experimental Evaluation}

\subsection{General setup} 

\subsubsection{Model Family \& Forget Set}
 
We use the GPT-Neo model family with 125 million (M), 1.3 billion (B), and 2.7 billion (B) parameters since: 1) they are known to emit verbatim responses and 2) they are the only large-enough language models used in previous research on unlearning for copyright purposes \cite{kassem2023preserving, jang2022knowledge}. 
Thus, with the GPT-Neo family a fair comparison with previous work can be secured.

The textual sentences we unlearn from the models (i.e., set $D_f$) are picked arbitrarily from the Training Data Extraction Challenge, with samples publicly available at \cite{gitjang}. These samples contain 200 tokens each, and they are easy-to-extract sequences from the Pile Corpora \cite{gao2020pile}, the dataset used to pretrain the GPT-Neo family. 

\subsubsection{Baselines}

\begin{figure*}[ht]
\vskip 0.2in
\begin{center}
\centerline{
  \includegraphics[width=0.33\linewidth]{ 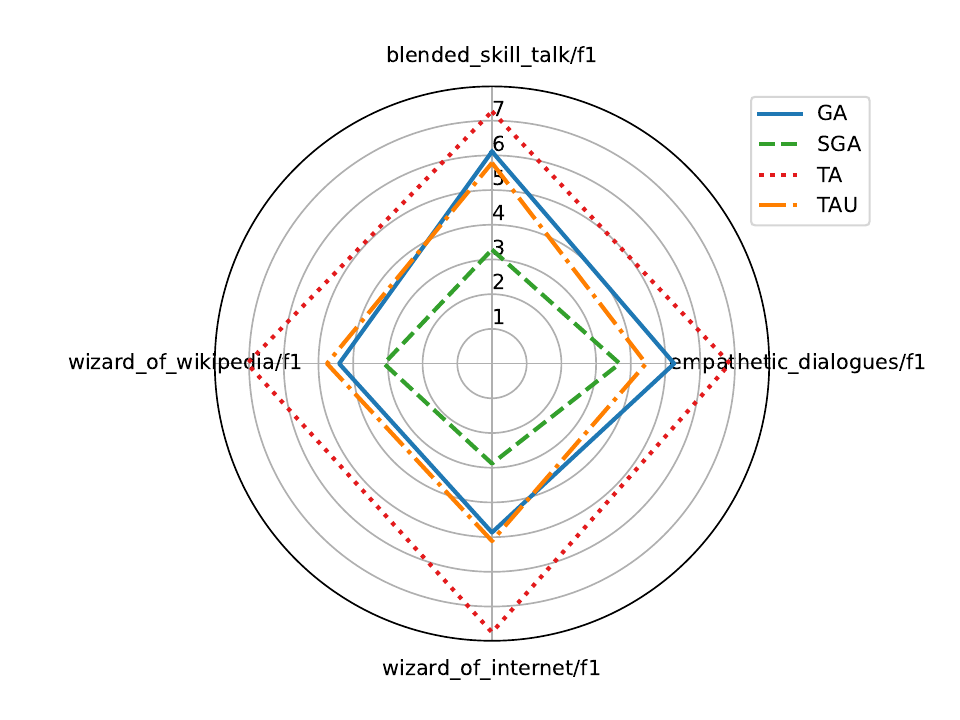}
  \includegraphics[width=0.33\linewidth]{ 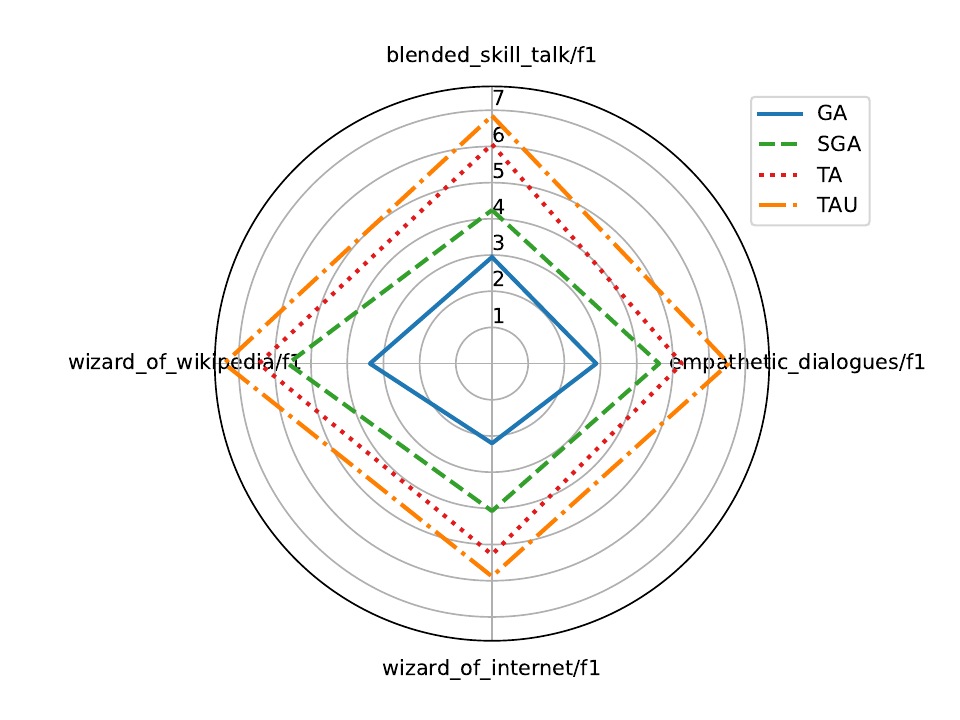}
  \includegraphics[width=0.33\linewidth]{ 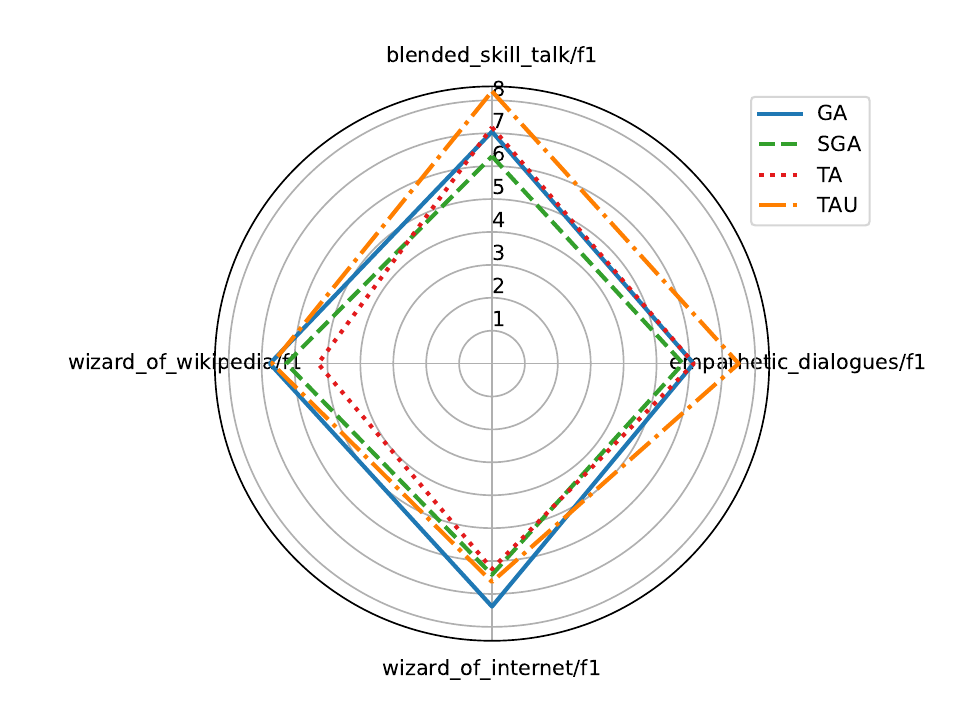}
}
\vspace{-0.5cm}

\caption{Average F1 scores for dialogue tasks. Unlearning 8 (left), 32 (middle), and 128 (right) samples. On the 125M model. The key discovery here is that TA and/or TAU perform best in small LLMs.
}
\label{dial_fsize}
\end{center}
\vskip -0.2in
\end{figure*}

We use SOTA unlearning methods in LLMs for unlearning memorized examples in the context of language generation. Given our taxonomy from \cref{related}, GA \cite{jang2022knowledge} is the SOTA. We also use TA \cite{ilharco2022editing} - by adapting it for unlearning training text from LLMs. We also consider differential privacy decoding (DPD) \cite{majmudar2022differentially} to shed light in utility-privacy trade-offs in text generation tasks. 

Our hardware setup and the hyperparameters are reported in \cref{hardware}.
\subsubsection{Evaluation Metrics}

To evaluate privacy, we report the number of `extractable' outliers after applying each algorithm, computed as per \cref{counteq} with $lb=15\%$ and $ub=60\%$. We stop unlearning when \cref{ineq} becomes true or if the algorithm stops making progress. Given generally, our privacy objective is to mimic the average memorization score of an unseen population of textual sequences $\frac{1}{|D_f|}\sum_{x_f \in D_f}g(x_f) \approx \frac{1}{|D_t|} \sum_{x_t \in D_t}g(x_t)$; we report the distance between the two averages at the end. 

To study the privacy-utility trade-offs, we use the target LLMs after unlearning to generate text in four popular benchmarks: Wizard of Wikipedia (WoW)\cite{dinan2018wizard}, Wizard of Internet (WoI) \cite{komeili2021internet}, Blended Skill Talk (BST) \cite{smith2020can}, and Empathetic Dialogues (ED) \cite{rashkin2018towards}.
We compute the F1 score (harmonic mean between precision and recall) and report the cross entropy (CE) with respect to the ground truth continuation for each dialogue. 
 
Furthermore, we evaluate the target model for common sense and scientific reasoning post-unlearning by using a suite of popular NLP classification tasks: ARC-Easy and ARC-Challenge \cite{clark2018think}, COPA \cite{gordon2012semeval}, Lambada
\cite{paperno2016lambada}, Winogrande \cite{sakaguchi2021winogrande}, PubmedQA \cite{jin2019pubmedqa}, Piqa \cite{bisk2020piqa}, Hellaswag \cite{zellers2019hellaswag}, and MathQA \cite{amini2019mathqa}. Across all the classification tasks, we report the average accuracy. 

\subsection{Performance Overview}

\textbf{Experiment setup for \cref{metrics_table}.} 
Five forget sets $D_f^i$ where $ i \in [0,4]$, are constructed with 16 textual sequences, $\|D_f^i\| = 16$, sampled from the Pile corpora. Each unlearning algorithm operates on a forget set. 
For each algorithm the results are averaged over the 5 forget sets. At the end, we report the mean across the average results for the dialogue tasks (F1, cross-entropy (CE)), and the classification tasks (ACC). We also report $\Delta$\texttt{MEM} as the difference between the memorization threshold and the memorization average at the end, to uncover outliers (but do not use it to report privacy quality). Instead, we rely on the number of outliers (EXTRACTABLE) to quantify protection. 

\textbf{Key insights.} We unveil six core insights. 
1) GA and SGA destructively ensure privacy in small LLMs,  sacrificing $80\%$  of the language generation capabilities in the 125M parameter model. 
In contrast, the task arithmetic-based methods (TA, TAU) reduce the F1 score across the four dialogue tasks by $30\%$. The destructive nature of GA in small models has been independently recognised in previous work \cite{jang2022knowledge} (the work that proposed GA). Uncovering reliable unlearning solutions for small-to-medium-sized  LLMs is paramount to enabling language modelling on hardware with limited computational resources. 
2) Larger models are more robust to the process of unlearning. Even simple methods such as GA perform reasonably well when the number of tunable parameters is high. We highlight that our algorithms (TAU and SGA) ensure higher utility across all tasks than the baselines. For our algorithms, we remark that the average F1 score and the cross-entropy are aligned with the default model (i.e., the model before unlearning). For the 2.7B model, we observe TAU surpasses the language modelling capabilities of the original model. 3) The strongest privacy guarantees are brought by SGA across all experiments, with a maximum of 1.8 samples extractable under our memorization thresholds. While the thresholds are set empirically, this gives us a good intuition on whether the individual memorization scores across the individuals in the forget set are aligned with the forgetting threshold. 
4) In contrast to the large drops in the ability to generate text, the model's reasoning abilities (i.e., the average accuracy across the classification tasks) is more robust to unlearning (results in \cref{app_class}). This is true for all model sizes. 
5) Differential private decoding (DPD) and Task Arithmetic (TA) have a good privacy-to-utility trade-off in small models (125M) but fail when the number of parameters exceeds 1B. This can also be observed by comparing the average loss when moving from small to large models, in the case of TA. 
6) On average, SGA and TAU incur a higher runtime overhead than GA and TA. Nonetheless, the runtime penalty is insignificant when putting it in the context of retraining from scratch or training with DP learning, and given the high utility and privacy benefits.

\cref{app_f1} has an in-depth evaluation of the language generation capabilities varying forget-set and model sizes.

\begin{figure*}[ht]
\vskip 0.2in
\begin{center}
\centerline{
  \includegraphics[width=0.43\linewidth]{ 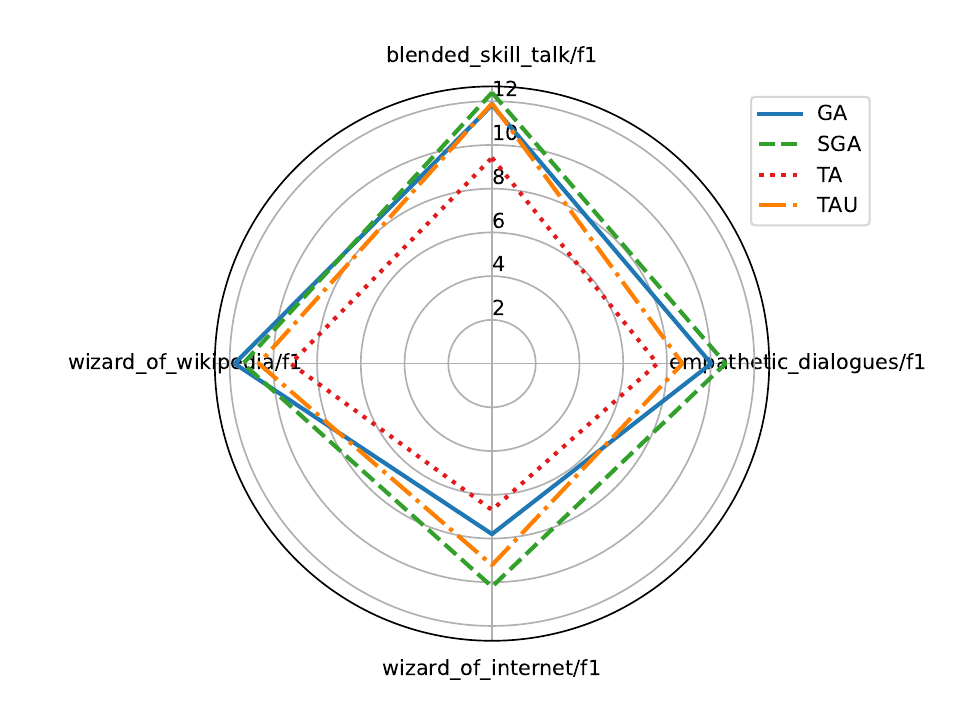}
  \includegraphics[width=0.43\linewidth]{ 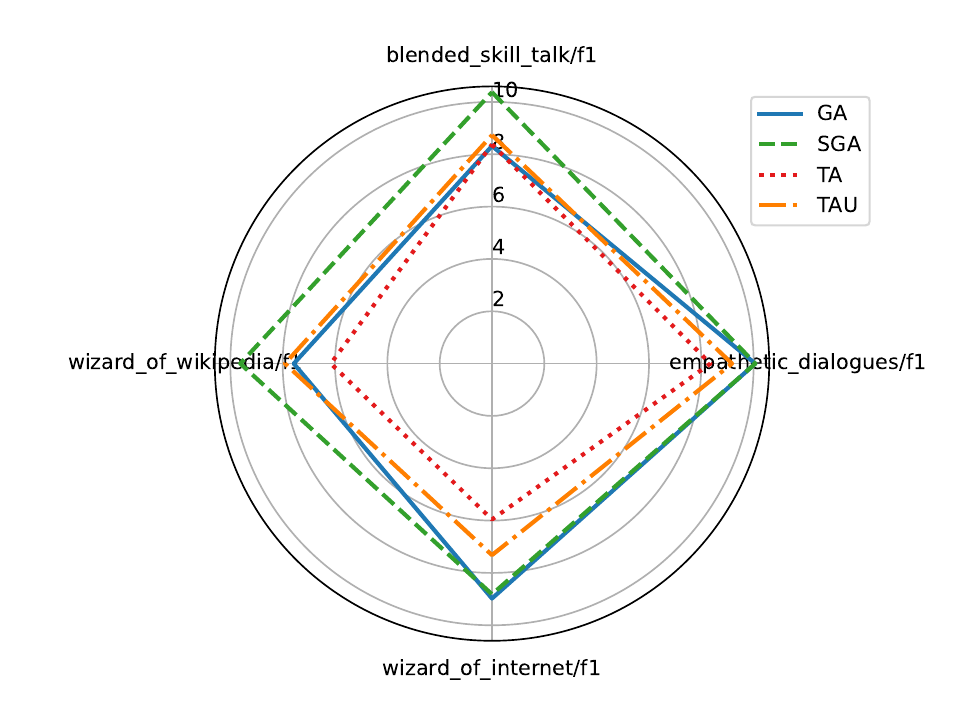}
}
\vspace{-0.5cm}
\caption{Average F1 scores for dialogue tasks. Unlearning 32 samples. On the 1.3B (left) and 2.7B (right) models. The key discovery here is that SGA catches up and surpasses the rest for larger models. 
}
\label{dial_param}
\end{center}
\vskip -0.2in
\end{figure*}

\subsection{Text Generation vs Forget Set Size}

\textbf{Experiment setup for \cref{dial_fsize}}. 
We construct 15 forget sets, 5 for each cardinality from $\{8,32,128\}$. 
Text generation after unlearning is tested on the four dialogue tasks. 

\textbf{Key insights.} There are three overarching findings concerning the trade-off between text generation and the size of the forget set. 1) Generally, task arithmetic (TA), and task arithmetic for unlearning (TAU) perform best across different forget-set cardinalities. Task arithmetic performs best when the number of samples is relatively small (we refer to Table \ref{metrics_table} for the results on 16 samples), and we observe from the left-most spider plot in \cref{dial_fsize} it greatly surpasses the other baselines when the number of samples forgotten at once is 8. TAU outperforms TA when the number of samples increases. We observe TAU outperforms the baselines if 32 and 128 samples, respectively, are forgotten at once. 2) Notably, gradient-based unlearning algorithms improve their trade-off between privacy and utility when the forget set size increases. On average, GA surpasses TAU when the cardinality of the forget set is 128, in the BST task. 3) There is a steep contrast in the ability to generate text between unlearning methods. When the forget set is small, gradient-based methods lose ca. 80\% in F1 for the 125M vs the 1.3B and 2.7B models (see \cref{metrics_table}).

\subsection{Text Generation vs Number of Model Parameters}

\textbf{Experiment setup for \cref{dial_param}.} 
We investigate unlearning methods w.r.t. their ability to generate text with different model sizes. We reuse the forget sets of 32 samples from above and unlearn these from the 1.3B and 2.7B GPT-Neo models. \cref{metrics_table} has results for forget sets of 16 samples.

\textbf{Key insights.} There are 2 key findings: 1) SGA outperforms competitors in text generation when the number of parameters exceeds 1B. This is in contrast to our findings with small models (see \cref{dial_fsize}), where gradient-based methods underperform regardless of the forget set size.  
Looking at F1 scores at task level (e.g., empathetic dialogues, etc.) we conclude that SGA sacrifices less utility while ensuring the strongest privacy guarantee, based on  \cref{counteq}). For instance, when the forget sets have a cardinality of 16, only 1 textual sample is found outside the interval $(15\%, 60\%)$. The same results are seen in big models as we vary the forget set size (see \cref{app_extended}). 2) In contrast to our findings from \cref{metrics_table}, increasing the size of the forget set in large models (from 16 to 32 here), TAU has a worse text generation utility-privacy trade-off than SGA. 

\section{Lessons Learned}

To our knowledge, this is the first work to contribute the following findings: 
1) The forget set may indeed contain extractable data points after applying various types of promising unlearning algorithms (GA, TA, DPD). We designed a new MIA against GA to show this (for space reasons, results are in \cref{streissand}).
2) There exist unlearning algorithms that innately minimize the number of forget-set outliers (e.g., SGA), which performs excellently when the model and the forget-set size scale up. 
3) While TA, as proposed by \cite{ilharco2022editing}, does not scale well 
to large forget-set and model sizes, it is a strong performer for smaller models. This fills an important gap as SOTA unlearning solutions so far sacrifice the model's utility in smaller models, hindering pushing LMs to devices with limited resources. 
4)  
TAU becomes on par with SGA and even exceeds its performance when the model is large and forget sets are smaller.  
5) There exists a 
unlearning algorithm (TAU) that has a 3X performance improvement in text generation than previous SOTA. 
6) After unlearning, the LLMs' reasoning abilities appear to degrade much less than the ability to generate text. This is an important result, which warrants more research.

\section{Limitations}
First, while we discover new unlearning algorithms with SOTA performance for unlearning memorized data from LLMs, we also unveil there is no single solution that performs best across the entirety of our problem space (i.e. different forget set sizes and model sizes). This finding is a consistent limitation in the Machine Unlearning research community \cite{kurmanji2023towards, kurmanji2024machine, maini2024tofu}. Second, our research aims to pave the way for efficient unlearning of textual sequences. To this end, we quantify privacy as the propensity of the model to regurgitate the training data verbatim (a.k.a. memorization). An adjacent problem setting to ours is unlearning factual associations from the model. In this line of work, an attacker may try to extract the link between two entities (e.g. PII data associated with a particular individual from the pre-training corpora). While we do our best to show empirically an unlearnt textual sequence (containing the PII information) will not be revealed verbatim by the LLM, a robust auditing process must be in place to ensure there are no further links between entities and private information. We deem our work as one of the first steps towards removing private information, and higher-level problems, such as eroding associations between entities, benefit from the initial steps we make in this direction.

\section{Conclusion}
We explored the problem of unlearning memorized data in LLMs from a fresh perspective: Each textual
sequence to be forgotten is treated differently, based on its memorization score. 
We adopted the TA method (for detoxifying LLMs) and showed that it can be SOTA in certain important cases. 
We proposed two new methods: SGA and TAU, which introduce the new unlearning perspective into GA and TA.
We devised a new MIA that shows that SOTA methods (like GA) that do not adopt this perspective are vulnerable to privacy attacks.
We demonstrated clear performance gains of these methods for model utility and for our new metric (for quantifying success for privacy and copyright reasons)  across an extensive suite of NLP tasks. This testifies to the importance of this fresh perspective.

\section{Broader Impact}
Recent advancements in deep learning and LLMs in particular have arrived at a fast pace, but come with increased responsibility: As researchers, we need to ensure the unintended behaviours of deep learning models are addressed thoroughly, be they harmful biases, privacy or copyright violations etc. Our research aims to take a fundamental step towards alleviating a class of privacy concerns from LLMs. For similar reasons, we do not store, process, or report any qualitative data related to the Pile corpora \cite{gao2020pile}, and only use the publicly available samples at \cite{gitjang}.

\bibliography{example_paper}
\bibliographystyle{icml2024}

\newpage
\appendix
\onecolumn

\section{Accuracy for Text Generation Results} 
\label{app_f1}

\begin{table}[ht]
\caption{Average F1 scores for the dialogue tasks: WoI, BST, ED, and WoW. Averages computed across five forget sets. 
TA is better for small models (125M). As forget-set sizes increase, TAU catches up to TA and surpasses it for larger models. SGA is weaker for smaller models and smaller forget sets. SGA catches up and surpasses GA as models and/or forgets sets get larger, where it becomes the winner overall. Overall, one or more of the new methods examined here, namely TA, TAU, and SGA, is the best performer.
}
\label{f1_full_table}
\vskip 0.15in
\begin{center}
\begin{small}
\begin{sc}
\begin{tabular}{lcccccccc}
\toprule
\multirow{2}{*}{Model} & \multirow{2}{*}{Algo} & \multirow{2}{*}{WoI/F1 $\uparrow$} & \multirow{2}{*}{BST/F1 $\uparrow$} & \multirow{2}{*}{ED/F1 $\uparrow$} & \multirow{2}{*}{WoW/F1 $\uparrow$} & \multirow{2}{*}{AVG F1 $\uparrow$} & \multirow{2}{*}{Forget Size} \\
& & & & & & & \\
\midrule
GPT-Neo(125M) & GA  & 5.25  & 6.11  & 4.87  & 4.40  & 5.16 & \multirow{12}{*}{8}\\
& SGA & 3.68  & 3.28  & 2.88  & 3.11  & 3.24  \\
& TA  & 6.85  & 7.27  & 7.74  & 7.05  & 7.23  \\
& TAU & 4.44  & 5.78  & 5.11  & 4.75  & 5.02  \\
\cmidrule{1-7}
GPT-Neo(1.3B)  & GA  & 11.83  & 10.06  & 10.81  & 12.70  & 11.35  \\
& SGA & 12.52  & 11.51  & 12.21 & 12.85  & 12.27  \\
& TA  & 9.28   & 6.69   & 6.10   & 9.52   & 7.90   \\
& TAU & 11.81  & 10.48  & 10.52  & 12.24  & 11.26  \\
\cmidrule{1-7}
GPT-Neo(2.7B) & GA  & 9.42  & 18.25  & 9.92  & 9.79  & 11.84  \\
 & SGA & 9.54  & 18.71  & 8.39  & 7.84  & 11.12  \\
 & TA  & 5.77  & 14.56  & 6.43  & 7.43  & 8.55   \\
 & TAU & 7.74  & 19.18  & 9.74  & 8.22  & 11.22  \\
\midrule
GPT-Neo(125M) & GA  & 2.72 & 2.23 & 2.37 & 2.23 & 2.28 & \multirow{12}{*}{16} \\
& SGA & 2.77 & 0.90 & 1.71 & 2.06 & 1.86 & \\
& TA  & 7.95 & 6.24 & 8.10 & 6.43 & 7.18 & \\
& TAU & 9.27 & 6.26 & 8.68 & 5.29 & 7.37 & \\
\cmidrule{1-7} 
GPT-Neo(1.3B) & GA  & 13.17 & 10.01 & 10.33 & 11.83 & 11.34 & \\
& SGA & 13.45 & 12.08 & 10.94 & 14.03 & 12.63 & \\
& TA  & 10.90 & 7.48  & 8.56  & 11.46 & 9.60  & \\
& TAU & 13.50 & 11.67 & 10.69 & 14.21 & 12.52 & \\
\cmidrule{1-7} 
GPT-Neo(2.7B) & GA  & 8.67 & 14.56 & 10.15 & 10.59 & 10.99 & \\
& SGA & 10.04 & 14.29 & 10.05 & 9.38 & 10.94 & \\
& TA  & 7.26  & 13.72 & 7.81  & 11.34 & 10.03 & \\
& TAU & 10.39 & 14.57 & 9.35  & 11.03 & 11.34 & \\
\midrule
GPT-Neo(125M) & GA & 2.20  & 3.37  & 2.94  & 2.88  & 2.85 & \multirow{12}{*}{32}\\
& SGA & 4.07  & 5.64  & 4.24  & 4.62  & 4.64  \\
& TA  & 5.28  & 6.44  & 6.06  & 5.24  & 5.76  \\
& TAU & 5.88  & 7.40  & 6.85  & 6.54  & 6.67  \\
\cmidrule{1-7} 
GPT-Neo(1.3B) & GA  & 11.77 & 7.80 & 10.02 & 11.83 & 10.36 \\
& SGA & 11.38 & 10.20 & 10.68 & 12.39 & 11.17 \\
& TA  & 9.20  & 6.69  & 7.56  & 9.42  & 8.22 \\
& TAU & 10.67 & 9.21  & 8.71  & 11.90 & 10.12 \\
\cmidrule{1-7}
GPT-Neo(2.7B) & GA & 7.58  & 10.06  & 8.32  & 8.98  & 8.73  \\
& SGA & 9.62  & 10.06  & 10.37  & 8.82 & 9.72 \\
& TA  & 6.16  & 8.35  & 8.38  & 5.96  & 7.21  \\
& TAU & 7.91  & 9.17  & 8.73  & 7.32  & 8.28  \\
\bottomrule
\end{tabular}
\end{sc}
\end{small}
\end{center}
\end{table}

\newpage
\section{Extended Performance Evaluation}
\label{app_extended}

\begin{table*}[h]
\caption{Performance overview across the suite of unlearning algorithms when forgetting 8 samples at once (averaged over 5 forget sets). All algorithms perform comparably for CE and ACC. For F1, as before, in the vast majority of times, one of TA/TAU/SGA is the clear winner. For the other metrics there exist large differences. Note the EXTRACTABLE column: SGA outperforms all. And that TA/TAU outperform GA.}
\vskip 0.15in
\begin{center}
\begin{small}
\begin{sc}
\begin{tabular}{lccccccc}
\toprule
\multirow{2}{*}{Model} & \multirow{2}{*}{Algo} & \multirow{2}{*}{F1 $\uparrow$} & \multirow{2}{*}{CE $\downarrow$} & \multirow{2}{*}{ACC $\uparrow$} & \multirow{2}{*}{EXTRACTABLE $\downarrow$} & \multirow{2}{*}{RUNTIME $\downarrow$} & \multirow{2}{*}{$\Delta$MEM $\downarrow$} \\
& & & & & & & \\
\midrule
GPT-Neo(125M) 
& GA  & 5.16 & 5.31 & 38.85 & 2.4 & 9.14 & 1.22 \\
& SGA & 3.24 & 6.02 & 38.36 & 0.2 & 10.66 & 1.96 \\
& TA  & 7.23 & 4.66 & 40.10 & 1.0 & 8.12 & 0.97 \\
& TAU & 5.02 & 6.36 & 39.09 & 3.2 & 31.03 & 1.35 \\
\midrule
GPT-Neo(1.3B) 
& GA  & 11.35 & 3.50 & 50.83 & 3.0 & 15.02 & 4.04 \\
& SGA & 12.27 & 3.37 & 49.65 & 0.4 & 22.25 & 3.19 \\
& TA  & 7.90  & 3.85 & 46.25 & 1.0 & 18.12 & 1.31 \\
& TAU & 11.26 & 3.46 & 50.66 & 1.6 & 39.83 & 2.21 \\
\midrule
GPT-Neo(2.7B) 
& GA  & 11.84 & 3.55 & 51.11 & 3.6 & 23.62 & 4.86 \\
& SGA & 11.12 & 3.42 & 53.33 & 0.6 & 25.88 & 3.62 \\
& TA  & 8.55  & 4.19 & 43.34 & 1.0 & 19.93 & 4.03 \\
& TAU & 11.22 & 3.71 & 50.00 & 1.4 & 52.46 & 1.29 \\
\bottomrule
\end{tabular}
\end{sc}
\end{small}
\end{center}
\vskip -0.1in
\end{table*}

\begin{table*}[h]
\caption{Performance overview across the suite of unlearning algorithms when forgetting 32 samples at once (averaged over 5 forget sets). The conclusions are the same as in the results of Table 4.}
\vskip 0.15in
\begin{center}
\begin{small}
\begin{sc}
\begin{tabular}{lccccccc}
\toprule
\multirow{2}{*}{Model} & \multirow{2}{*}{Algo} & \multirow{2}{*}{F1 $\uparrow$} & \multirow{2}{*}{CE $\downarrow$} & \multirow{2}{*}{ACC $\uparrow$} & \multirow{2}{*}{EXTRACTABLE $\downarrow$} & \multirow{2}{*}{RUNTIME $\downarrow$} & \multirow{2}{*}{$\Delta$MEM $\downarrow$} \\
& & & & & & & \\
\midrule
GPT-Neo(125M) 
& GA  & 2.85 & 6.22 & 41.56 & 9.0 & 9.06 & 2.20 \\
& SGA & 4.64 & 5.60 & 39.48 & 3.2 & 12.22 & 0.79 \\
& TA  & 5.76 & 4.85 & 38.12 & 5.0 & 12.14 & 0.56 \\
& TAU & 6.67 & 4.56 & 42.22 & 6.2 & 71.28 & 7.63 \\

\midrule
GPT-Neo(1.3B)

& GA  & 10.36 & 3.60 & 51.46 & 9.4 & 19.82 & 3.93 \\
& SGA & 11.17 & 3.48 & 49.51 & 2.2 & 29.09 & 1.29 \\
& TA  & 8.22  & 4.00 & 44.37 & 2.6 & 19.69 & 1.10 \\
& TAU & 10.12 & 3.59 & 49.23 & 8.8 & 44.16 & 0.62 \\
\midrule
GPT-Neo(2.7B) 

& GA  & 8.73 & 3.70 & 53.95 & 11.6 & 23.10 & 5.36 \\
& SGA & 9.72 & 3.54 & 55.20 & 2.0 & 29.45 & 3.11 \\
& TA  & 7.21 & 4.32 & 44.09 & 3.2 & 27.33 & 3.15 \\
& TAU & 8.28 & 3.63 & 53.89 & 10.0 & 40.60 & 1.49 \\

\bottomrule
\end{tabular}
\end{sc}
\end{small}
\end{center}
\vskip -0.1in
\end{table*}

\section{Hyperparameters \& Hardware Settings}
\label{hardware}


\begin{table*}[t]
\caption{Hyperparameters}
\label{hyper-table}
\vskip 0.15in
\begin{center}
\begin{small}
\begin{sc}
\begin{tabular}{lcccrc}
\toprule
Algo & Hyperparameter & Symbol & Value \\
\midrule
DPD & linear interpolation factor  & $\lambda$ & 0.2 \\
\midrule
GA&  learning rate & $\alpha$ & 5e-5 \\
\midrule
SGA &  learning rate & $\alpha$ & 5e-5 \\
&  memorization upper limit & $\gamma$ & 70\% \\
&  top-k & k & 1 \\
\midrule
TA &  learning rate & $\alpha$ & 5e-5 \\
& drift scaling factor & $\lambda$ & 3 \\
\midrule
TAU &  learning rate & $\alpha$ & 5e-5 \\
&  memorization upper limit & $\gamma_u$ & 70\% \\
&  memorization to switch & $\gamma_s$ & 50\% \\
& drift scaling factor  & $\lambda$ & 1 \\
\bottomrule
\end{tabular}
\end{sc}
\end{small}
\end{center}
\vskip -0.1in
\end{table*}

The hyperparameters used in our experiments are reported in \cref{hyper-table}. Differential privacy decoding (DPD) uses a linear interpolation $\lambda = 0.2 $ that allows the decoder to achieve the desired \textit{exact memorization} thresholds (see \cref{sample-table} for the thresholds).
We use the same learning rate for gradient ascent (GA) as in the original paper \cite{jang2022knowledge}, given unlearning is performed over the same data distribution. For SGA we have a memorization upper limit of 70\%. Subsequently, the loss is maximised with respect to the most memorised sample $k=1$. For task arithmetic, we perform a grid search to find the size of the drift that can reduce the average memorization to the threshold $\lambda=3$, which is more aggressive than the unlearning proposal from the original paper \cite{ilharco2022editing}. For TAU, we perform SGA with the same hyperparameters as above until the memorization score becomes $Y_s=50\%$. Subsequently, we scale the drift by a factor of $\lambda=1$ during task arithmetic for the last part of the unlearning process. 

In terms of hardware, we use two NVIDIA RTX A5000 with 24GB across all experiments. 

\section{Comparing against GA+Mismatch for Unlearning Memorized Data in LLMs}
In \cite{yao2024} the GA algorithm (as inspired by \cite{thudi2022unrolling} and a variation of GA, coined GA+Mismatch, were presented and studied as methods for unlearning in LLMs. GA and GA-Mismatch were shown to be top performers against finetuning. As the proposed and studied methods in our work (namely, TA, TAU, and SGA) outperform GA \cite{jang2022knowledge}, we expected to see similar behaviour against GA+Mismatch. To substantiate this, we conducted experiments using GPT-NEO(1.3B) with the results presented and summarized in Table \ref{new_results_table_mis}, where it is shown that GA+Mismatch (or MIS as we call it for brevity) lacks behind the new methods proposed here across all metrics.

\begin{table*}[ht]
\caption{Comparing GA+Mismatch \cite{yao2024} (MIS) against GA, SGA, TA, and TAU in GPT-NEO(1.3B). We see that MIS is never the top performer and most often a bad performer. Specifically: For F1, it is either the 4th or 5th best (worst). For CE and ACC it is always the 5th best (worst). For EXTRACTABLE it is either 2nd- or 3rd-best. In addition, MIS is always worse than GA for F1, CE, and ACC. Notably, however, MIS outperforms GA for EXTRACTABLE, which, in hindsight, is to be expected as the added noise by the mismatch term in the loss function of MIS tends to protect privacy.}
\label{new_results_table_mis}
\vskip 0.15in
\begin{center}
\begin{small}
\begin{sc}
\begin{tabular}{lccccccc}
\toprule
\multirow{2}{*}{GPT-Neo(1.3B)} & \multirow{2}{*}{Algo} & \multirow{2}{*}{F1 $\uparrow$} & \multirow{2}{*}{CE $\downarrow$} & \multirow{2}{*}{ACC $\uparrow$} & \multirow{2}{*}{EXTRACTABLE $\downarrow$} & \multirow{2}{*}{$\Delta$MEM $\downarrow$} \\
& & & & & & & \\
\midrule 

8 samples 
& GA  & 11.35 & 3.50 & 50.83 & 3.0 & 4.04 \\
& SGA & 12.27 & 3.37 & 49.65 & 0.4& 3.19 \\
& TA  & 7.90  & 3.85 & 46.25 & 1.0& 1.31 \\
& TAU & 11.26 & 3.46 & 50.66 & 1.6 & 2.21 \\
& MIS & 8.51 & 4.03 & 43.23 & 0.8 &3.15 \\

\midrule
16 samples 
        & GA  & 11.34 & 3.52 & 49.37 & 4.6  & 4.06 \\
         & SGA & 12.63 & 3.37 & 49.30  & 1.0 & 1.86 \\
         & TA  & 9.60  & 3.69 & 48.75 & 2.0& 1.80 \\
         & TAU & 12.52 & 3.39 & 49.24 & 4.0  & 0.81 \\
         & MIS & 8.99 & 4.02 & 42.15 & 3.6 & 0.88 \\
\midrule
32 samples

& GA  & 10.36 & 3.60 & 51.46 & 9.4  & 3.93 \\
& SGA & 11.17 & 3.48 & 49.51 & 2.2 & 1.29 \\
& TA  & 8.22  & 4.00 & 44.37 & 2.6  & 1.10 \\
& TAU & 10.12 & 3.59 & 49.23 & 8.8  & 0.62 \\
& MIS & 8.83 & 4.03 & 41.74 & 2.4  & 4.44 \\

\bottomrule
\end{tabular}
\end{sc}
\end{small}
\end{center}
\vskip -0.1in
\end{table*}

\section{The Effect of Parameter-Efficient Methods on Unlearning Memorized Data in LLMs}

New work on parameter-efficient (finetuning) methods (a.k.a. PEFT) \cite{peft2023} has shown that PEFT methods can help to detoxify LLMs. The studied methods there were LoRA and a new method proposed, called $(IA)^3$. As LoRA is shown to outperform $(IA)^3$, we conducted experiments to show how LoRA would affect the unlearning performance of GA, SGA, and TA in our problem setting. The results are shown and summarized in Table \ref{lora}.

\begin{table*}[h]
\caption{Investigating the effect of a Parameter Efficient FineTuning (PEFT) method \cite{peft2023} (LoRA) on unlearning memorized data in LLMs. The results show the effect of adding LoRA to GA, SGA, and TA versus GPT-NEO(125M). PEFT methods like LoRA cannot be viewed as unlearning algorithms per se. Nonetheless, they do affect the efficacy of unlearning. The results show that adding LoRA to (i) TA never improves TA for F1, ACC, CE, and EXTRACTABLE; 
(ii) SGA only benefits F1, but not ACC, CE, EXTRACTABLE or $\Delta$\texttt{MEM}; (iii) GA benefits for F1, CE, and EXTRACTABLE, but not for ACC or $\Delta$\texttt{MEM}. As expected, LoRA can indeed improve significantly the runtime of algorithms: by ca. 2X for GA and SGA. For TA the improvement is much smaller. In general, the big picture is expected to be nuanced and a separate comprehensive investigation should be conducted to assess the impact of PEFT methods on unlearning algorithms.}
\label{lora}
\vskip 0.15in
\begin{center}
\begin{small}
\begin{sc}
\begin{tabular}{lccccccc}
\toprule
\multirow{2}{*}{Parameters} & \multirow{2}{*}{Algo} & \multirow{2}{*}{F1 $\uparrow$} & \multirow{2}{*}{CE $\downarrow$} & \multirow{2}{*}{ACC $\uparrow$} & \multirow{2}{*}{EXTRACTABLE $\downarrow$} & \multirow{2}{*}{RUNTIME $\downarrow$} & \multirow{2}{*}{$\Delta$MEM $\downarrow$} \\
& & & & & & & \\
\midrule
GPT-Neo(125M) 
& GA  & 2.85 & 6.22 & 41.56 & 9.0 & 9.06 & 2.20 \\
& SGA & 4.64 & 5.60 & 39.48 & 3.2 & 12.22 & 0.79 \\
& TA  & 5.76 & 4.85 & 38.12 & 5.0 & 12.14 & 0.56 \\

\midrule
LoRA (r=32, $\alpha$=64) 
& GA+LoRA & 7.39 & 5.88 & 36.11 & 7.0 & 5.26 & 3.70 \\
& SGA+LoRA & 8.56 & 5.68 & 36.08 & 5.0 & 6.23 & 3.90 \\
& TA+LoRA & 4.68 & 5.78 & 33.99 & 5.0 & 13.07 & 0.69 \\

\bottomrule
\end{tabular}
\end{sc}
\end{small}
\end{center}
\vskip -0.1in
\end{table*}

\section{The Effect of Large Forget Sets}

\begin{table*}[h]
\caption{The effect of larger forget sets. We notice that for larger forget sets, the conclusions with respect to the utility metrics (i.e., F1, CE, and ACC) among GA, SGA, TA, and TAU remain as before. Namely, the SOTA GA is never the winner - at least one of the new solutions we propose outperforms GA. Similarly for the forget quality metrics i.e., EXTRACTABLE, where SGA and TA outperform GA.}
\label{large_forget}
\vskip 0.15in
\begin{center}
\begin{small}
\begin{sc}
\begin{tabular}{lccccccc}
\toprule
\multirow{2}{*}{Forget Set Size} & \multirow{2}{*}{Algo} & \multirow{2}{*}{F1 $\uparrow$} & \multirow{2}{*}{CE $\downarrow$} & \multirow{2}{*}{ACC $\uparrow$} & \multirow{2}{*}{EXTRACTABLE $\downarrow$} & \multirow{2}{*}{RUNTIME $\downarrow$} & \multirow{2}{*}{$\Delta$MEM $\downarrow$} \\
& & & & & & & \\
\midrule
256
& GA & 8.42 & 4.40 & 42.40 & 11.6 & 10.82 & 4.29 \\
& SGA & 8.59 & 4.33 & 42.36 & 8.4 & 11.06 & 2.91 \\
& TA & 8.01 & 4.20 & 42.53 & 3.8 & 35.67 & 0.73 \\
& TAU & 9.15 & 4.10 & 44.90 & 6.4 & 65.45 & 0.38 \\
\midrule
512
& GA & 9.02 & 4.24 & 44.10 & 9.4 & 11.35 & 2.61 \\
& SGA & 9.17 & 4.20 & 44.06 & 6.6 & 11.30 & 1.93 \\
& TA & 7.08 & 4.46 & 43.61 & 4.6 & 16.21 & 1.67 \\
& TAU & 8.48 & 4.18 & 44.58 & 10.8 & 20.98 & 1.59 \\

\bottomrule
\end{tabular}
\end{sc}
\end{small}
\end{center}
\vskip -0.1in
\end{table*}

\section{Adversarial Attacks for Textual Sequences}
\label{streissand}
Why do textual sequences completely forgotten during unlearning raise privacy concerns? To answer this question, we revert our attention to membership inference attacks (MIAs), adversarial attacks aimed at extracting training data points. We set up a loss-based MIA for Unlearning in LMs, which we coin MIAU, inspired from \cite{mattern2023membership}.
\begin{definition}
     Let $x\in X$ be a textual sequence, where $X$ is the set of all possible textual sequences over the model's vocabulary $\mathcal{V}$. Then we define $\tilde{x}$ as a \textbf{neighbour} to $x$ if $\tilde{x}$ is constructed with a bidirectional encoder $g$ from the corrupted version of $\hat{x}$ ($x$ with a percentage of tokens $w_i$ masked). Formally, for each masked token in $\hat{x}$ replace it with $\tilde{w_i} = g(\mathcal{V}^i, x)$
\end{definition}
\begin{definition}
    Let $x=(w_1,w_2,..,w_T)$ be a textual sequence of $T$ tokens with $x\in X$, where $X$ is the set of all possible textual sequences over the model's vocabulary $\mathcal{V}$. Let $f_\theta$ be a neural LM with parameters $\theta$ optimised under the causal language modeling loss function $\mathcal{L}_\theta(x)=-\sum_{t=1}^{T-1} log f_\theta(w_t | w_{<t})$. Let $\tilde{x}_1, \tilde{x}_2,..., \tilde{x}_n$ be a set of $n$ neighbours to $x$. Then we define \textbf{MIA for Unlearning (MIAU)} as:
    \begin{equation}
    \label{miau}
        \left| \mathcal{L}_\theta(x) - \frac{1}{n} \sum_{i=1}^{n} \mathcal{L}_\theta(\tilde{x}_i) \right| > \gamma
    \end{equation}
\end{definition}

The intuition behind the decision law in \cref{miau} is simple. The distance between the loss of the original textual sequence to the average loss of its neighbours is higher than a privacy threshold $\gamma$ when the model has memorised the original sequence or when the sequence has been completely removed, as we expect better generalisation over its neighbours in the latter case. This is because we treat the neighbours as validation points or, in Machine Unlearning parlance, as a proxy for `retrain from scratch'. Note that the memorization score defined in \ref{memeq} is an approximation of the causal language modelling loss function, hence the reason we use loss and memorization interchangeably here.

\begin{figure*}[ht]
\vskip 0.2in
\begin{center}

\centerline{
  \includegraphics[width=0.49\linewidth]{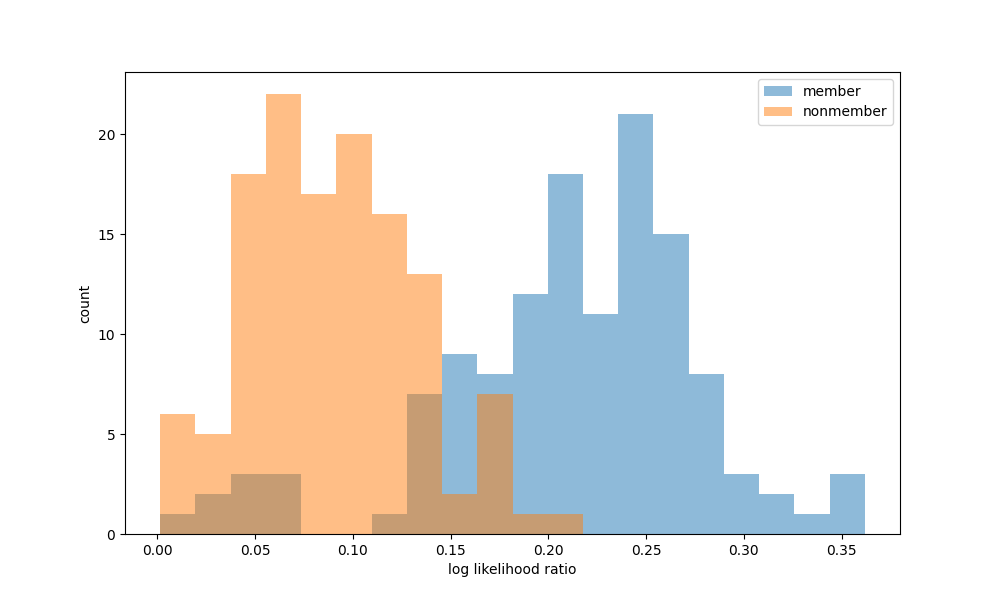}
  \includegraphics[width=0.49\linewidth]{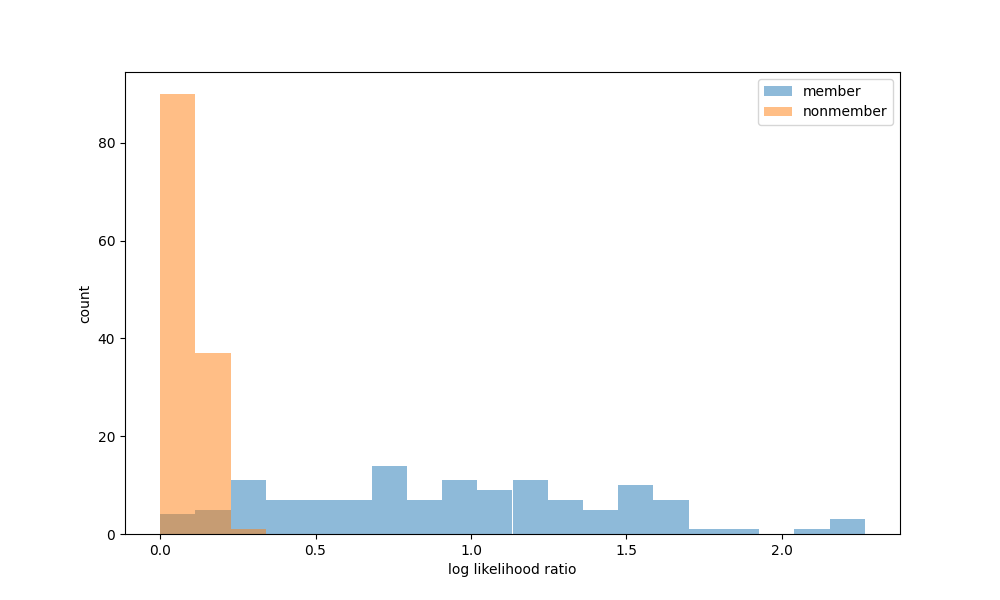}

}

\caption{From left to right: the loss distribution across 256 textual sequences by querying the original model (1.3B GPT-Neo checkpoint) and the model after unlearning. On the x-axis we see the distance of training samples (members) to their artificial neighbours (\cref{miau}). We use 128 samples from the enron email dataset \cite{shetty2004enron} as members and 128 samples from the SAMSum training dataset \cite{gliwa-etal-2019-samsum} as nonmembers. After unlearning the members with gradient ascent, they become easily extractable (less overlap with nonmembers) as the distance between their loss and that of their neighbours' becomes increasingly large.}
\label{MIA}
\end{center}

\vskip -0.2in
\end{figure*}

\begin{figure*}[ht]
\vskip 0.2in
\begin{center}

\centerline{
  \includegraphics[width=0.49\linewidth]{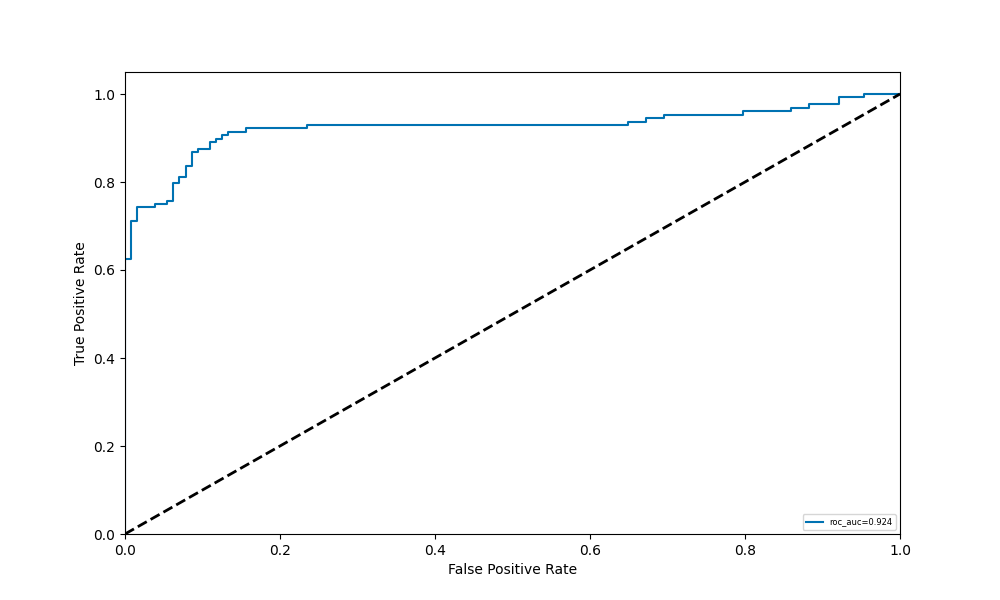}
  \includegraphics[width=0.49\linewidth]{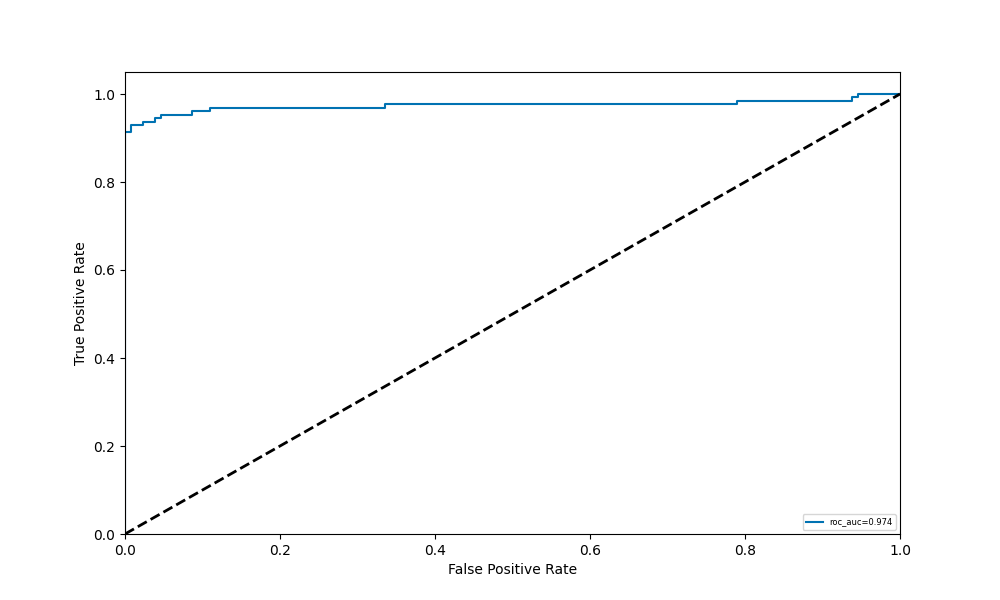}

}

\caption{From left to right: the ROC curves for the experiment setup in \cref{MIA} for the original model and the unlearned model. Remarkably, there is a $50\%$ increase in the success of the adversarial attack after unlearning, from a true positive rate (TPR) of $60\%$ to a TPR of $90\%$, at a false positive rate (FPR) of $0\%$. And there is a clear difference on FPR in the FNR region [0, 0.2], which includes the regions we should attend to, according to Carlini et al \cite{carlini2022membership}.}
\label{roc}
\end{center}

\vskip -0.2in
\end{figure*}

It is evident from \cref{MIA} and \cref{roc} that the members (i.e. the textual sequences) of the forget set become more extractable after unlearning if their memorization score is very low. This artifact of unlearning has been observed in previous Machine Unlearning work in computer vision and coined the ``Streisand effect" \cite{golatkar2020eternal}. We are the first, to the best of our knowledge, to prove it occurs in LLMs and to take a first step towards devising unlearning solutions that prevent it.

\newpage

\section{Unlearning Effectiveness via Loss-Distance Magnitude} \label{loss magnitude}

The membership inference strategy presented in \cref{streissand} is used to `count` the number of extractable points from the forget set after unlearning is applied. For a textual sequence to be qualified as a training or unlearnt `member', the absolute distance to its neighbours (\cref{miau}) needs to be higher than a certain appetite for false positive rate, which is generally very low. We note the same distance can be used as a measurement of model `indistinguishability' with respect to the forget set. We refer to model indistinguishability as the closeness in the behaviour of two models: (i) the unlearned model and (ii) models retrained from scratch without the forget set. In essence, the higher (lower) indistinguishability to retrain-from-scratch, implies that MIAs will be less (more) successful, leading to better (worse) privacy. 

To quantify model behaviour we inspect the model's loss, which is a principled approach and in the spirit of SOTA methods from privacy literature \cite{carlini2022membership}. To compare indistinguishability between two unlearning algorithms, A and B, we introduce the loss-distance magnitude (LOMA). We remark LOMA is independent of memorization (\cref{memeq}). LOMA is a fine-grained metric that tracks the distance to retrain-from-scratch by fixing the forget set and using the parameters at the end of the unlearning process in \cref{miau} (i.e. $\theta_A$ and $\theta_B$). Subsequently, we compare the two loss-distance distributions to the loss distances we would get for unseen data by the language model, which serves as our reference point for high indistinguishability (better privacy). A key finding is that feeding unseen data through \cref{miau} results in consistent distances oblivious to $\theta$. This is easily explained by the fact that unseen data had no impact on the parameters neither during pretraining nor unlearning.

\begin{figure*}[ht]
\vskip 0.2in
\begin{center}

\centerline{
  \includegraphics[width=0.49\linewidth]{ 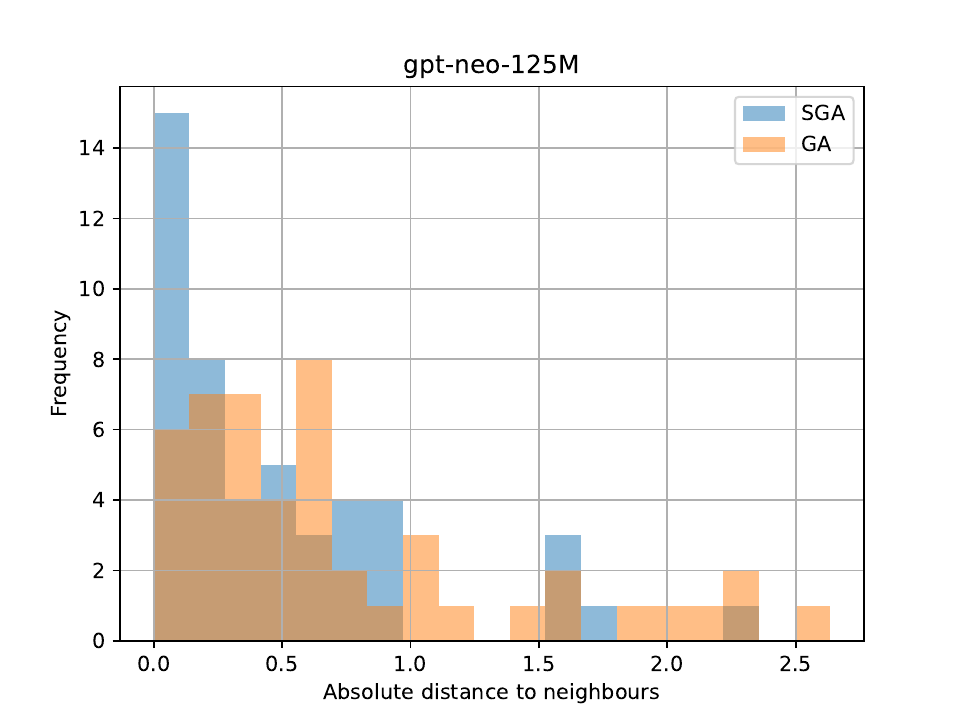}
  \includegraphics[width=0.49\linewidth]{ 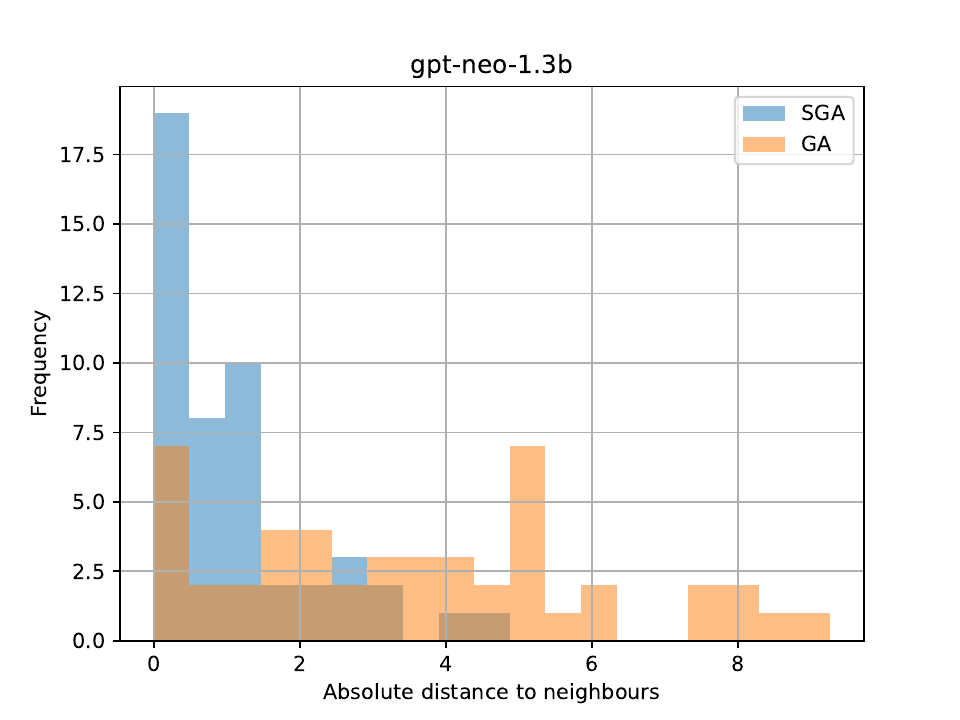}

}

\caption{We investigate the difference between the loss of our unlearnt model (with GA and SGA) and a retrain-from-scratch proxy (proxy `data neighbours' computed as shown in Appendix C). We do this for every sample in the forget set! Naturally, higher indistinguishability means much smaller distances (i.e., between 0 and 0.5 in our experiments on unseen data). The two figures show the loss distances of the forget set for the 125M and 1.3B checkpoints of GPT-Neo, aggregated across several forget sets/ unlearning runs.}
\label{LOMA_fig}
\end{center}

\vskip -0.2in
\end{figure*}

It becomes obvious from \cref{LOMA_fig} that SGA provides crystal clear privacy benefits. In the GPT-Neo (125M) we observe 23 samples in the forget set reside within [0, 0.25] with 15 samples in the first bin for SGA, compared to 6 for GA. Furthermore, we see GA has a long-tail distribution, with 13 sequences above a loss distance of 1, compared to only 5 for SGA. Similarly, the plot for 1.3B speaks for itself. There are 19 data points in the first bin [0, 0.5] for SGA and only 7 for GA. Furthermore, there are 16 examples with a loss distance higher or equal to 4 for GA, whereas there are zero such examples for SGA. Generally, the population of the unlearnt samples resides within small distances from the retrain-from-scratch proxy for SGA, while the distribution of distances for GA is much worse/flat with distances up to 9.5. While inspecting memorization can be used to extract data points (e.g. the points from the long-tail above), through this experiment we quantify the `magnitude` of extractability and the distance in model behaviour from the optimal unlearnt model (i.e. retrain-from-scratch).

\newpage

\section{Language Reasoning (Classification) Results} 
\label{app_class}

\begin{figure*}[ht]
\vskip 0.2in
\begin{center}
\centerline{
  \includegraphics[width=0.33\linewidth]{ 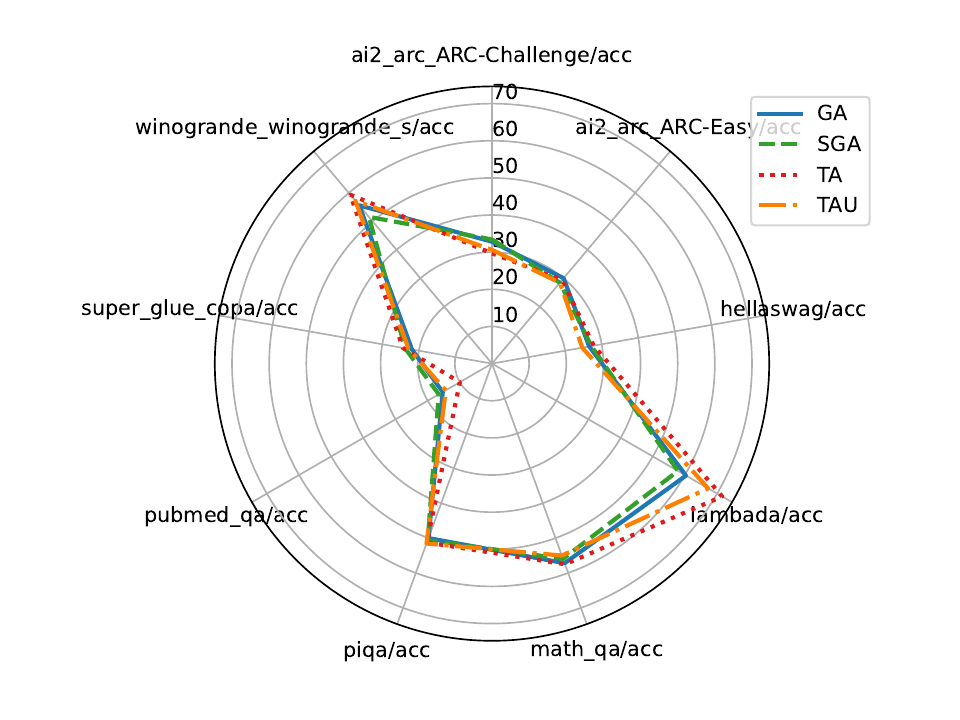}
  \includegraphics[width=0.33\linewidth]{ 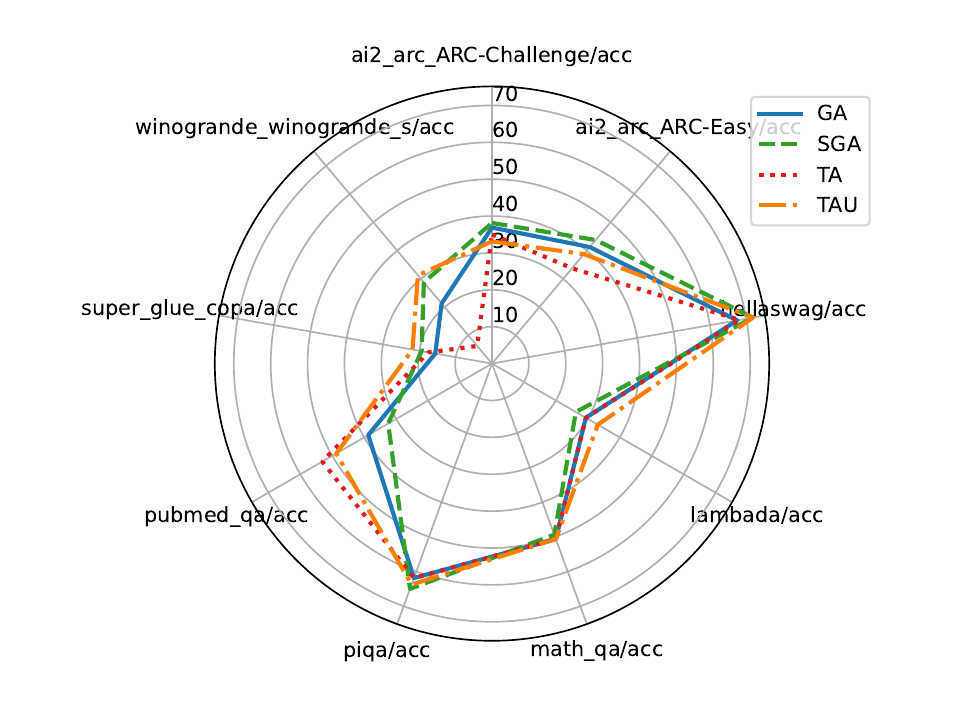}
  \includegraphics[width=0.33\linewidth]{ 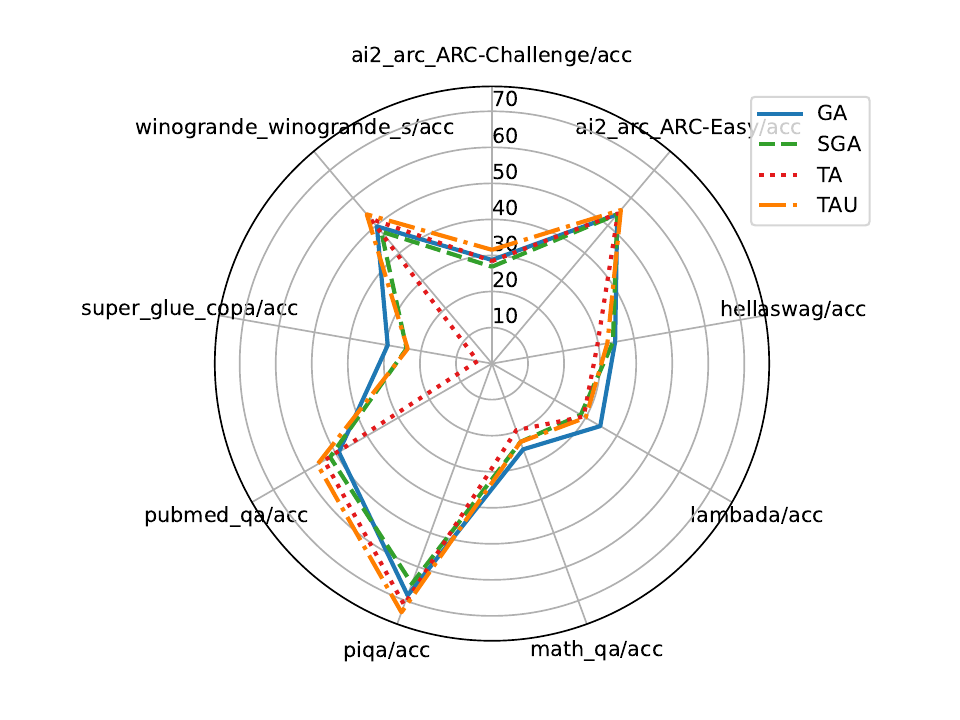}
}

\caption{The average accuracy scores across nine language reasoning tasks. From left to right: unlearning 8 samples, 16 samples, and 32 samples at once, respectively. Results based on the 125 million checkpoint. }
\end{center}
\vskip -0.2in
\end{figure*}

\begin{figure*}[ht]
\vskip 0.2in
\begin{center}
\centerline{
  \includegraphics[width=0.33\linewidth]{ 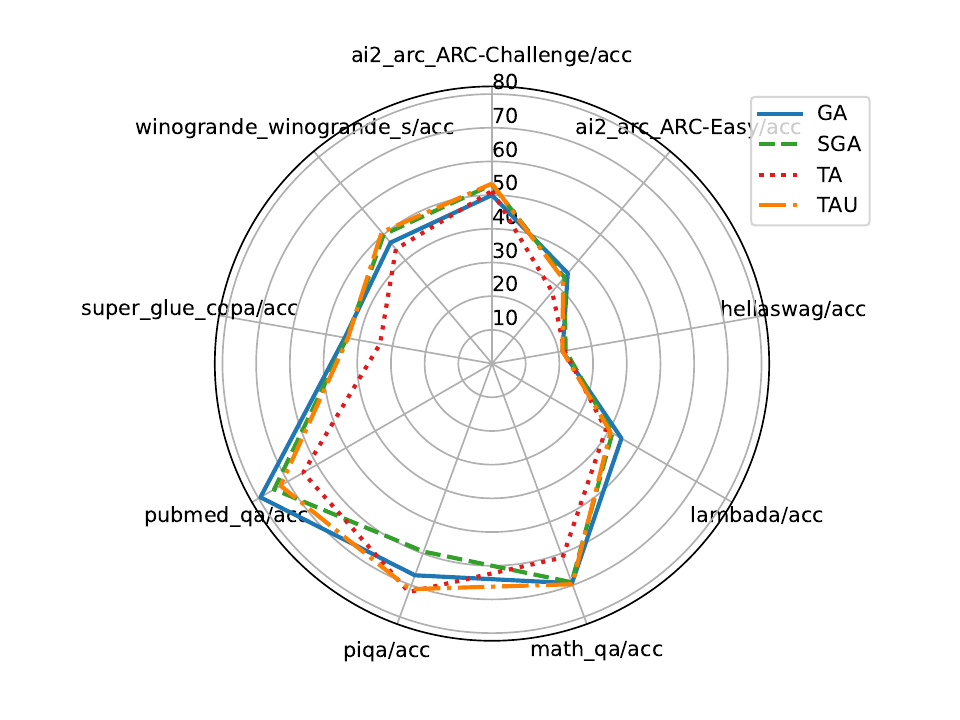}
  \includegraphics[width=0.33\linewidth]{ 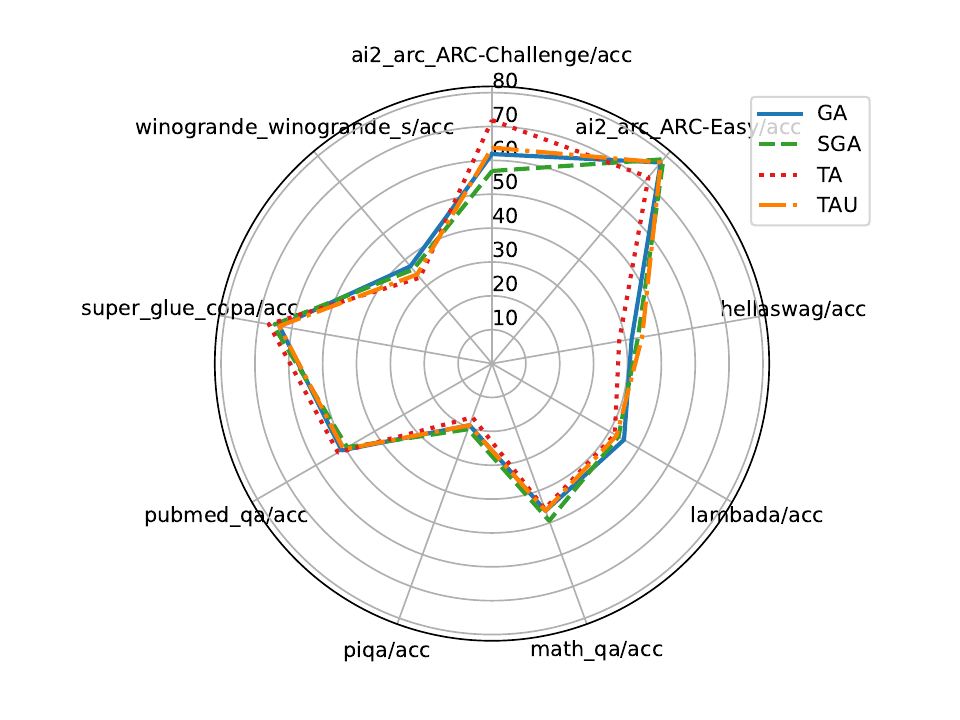}
  \includegraphics[width=0.33\linewidth]{ 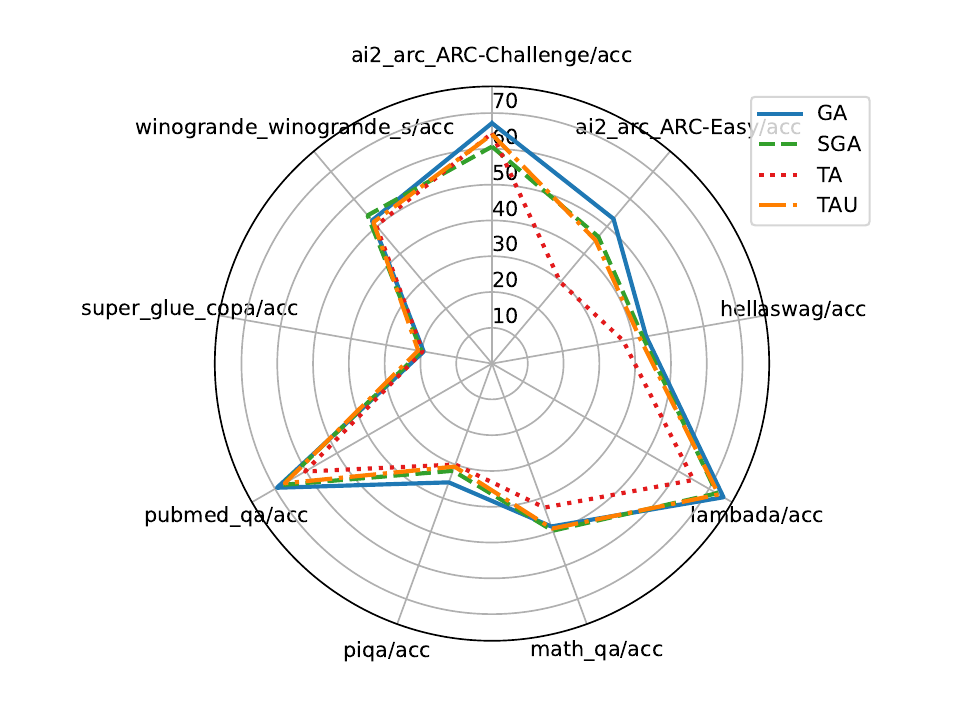}
}

\caption{The average accuracy scores across nine language reasoning tasks. From left to right: unlearning 8 samples, 16 samples, and 32 samples at once, respectively. Results based on the 1.3B checkpoint. }
\end{center}
\vskip -0.2in
\end{figure*}

\pagebreak

\begin{figure*}[ht]
\vskip 0.2in
\begin{center}
\centerline{
  \includegraphics[width=0.33\linewidth]{ 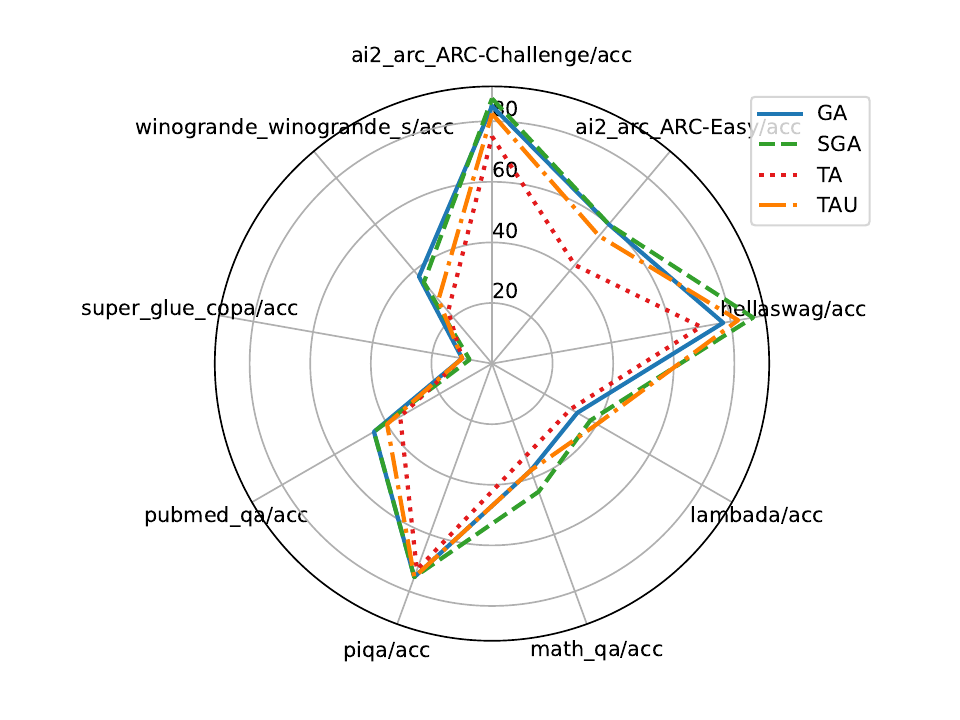}
  \includegraphics[width=0.33\linewidth]{ 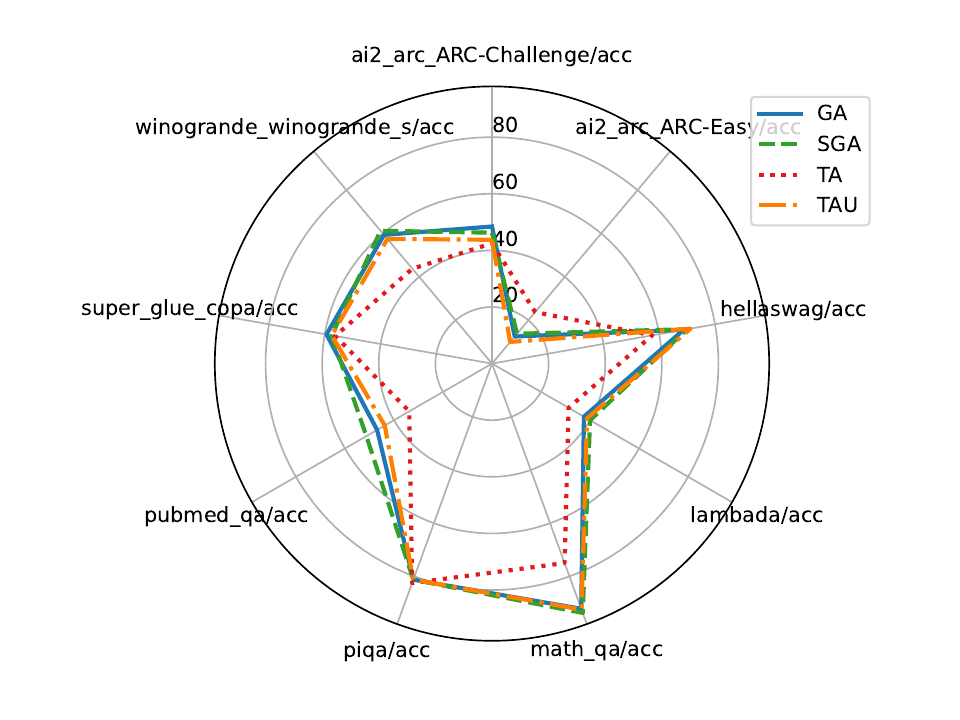}
  \includegraphics[width=0.33\linewidth]{ 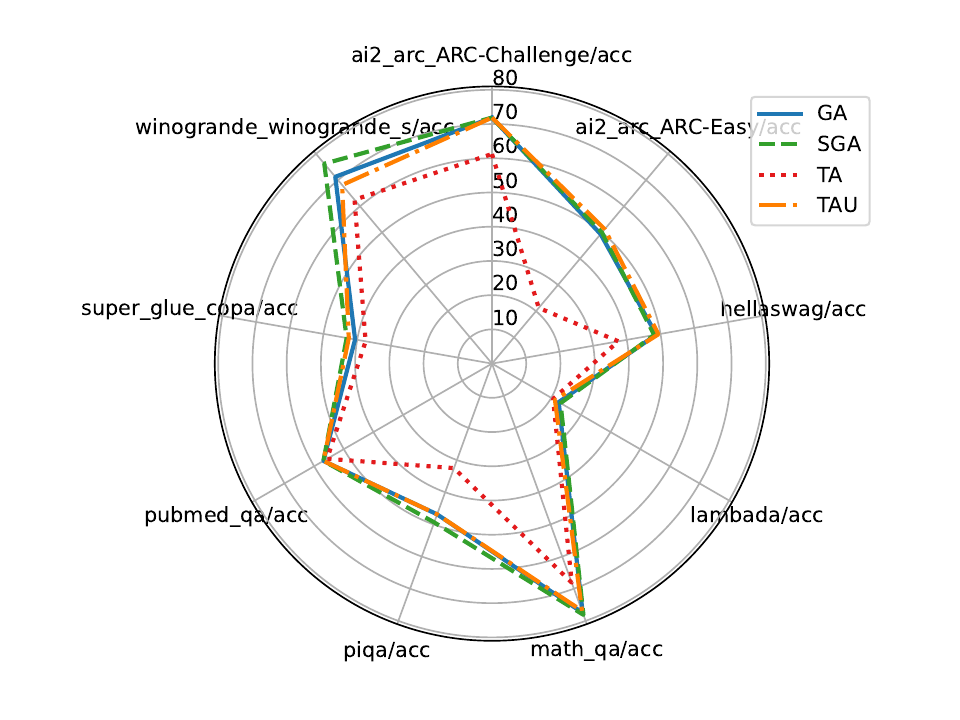}
}

\caption{The average accuracy scores across nine language reasoning tasks. From left to right: unlearning 8 samples, 16 samples, and 32 samples at once, respectively. Results based on the 1.3B checkpoint. }
\end{center}
\vskip -0.2in
\end{figure*}

\section{Loss on Dialogue Tasks}

\begin{figure*}[ht]
\vskip 0.2in
\begin{center}
\centerline{
  \includegraphics[width=0.33\linewidth]{ 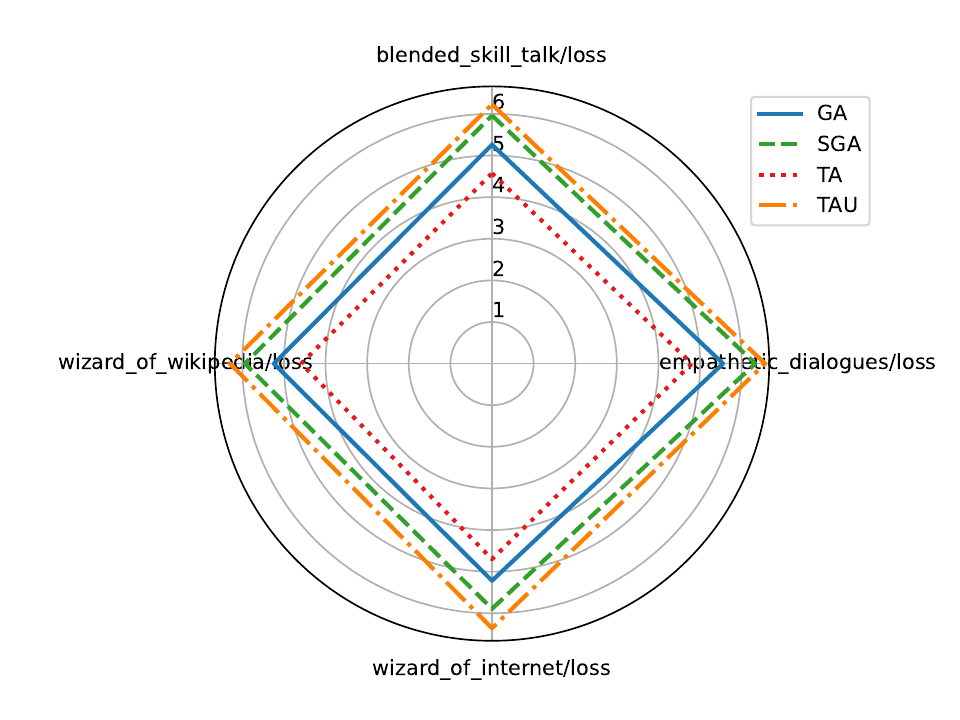}
  \includegraphics[width=0.33\linewidth]{ 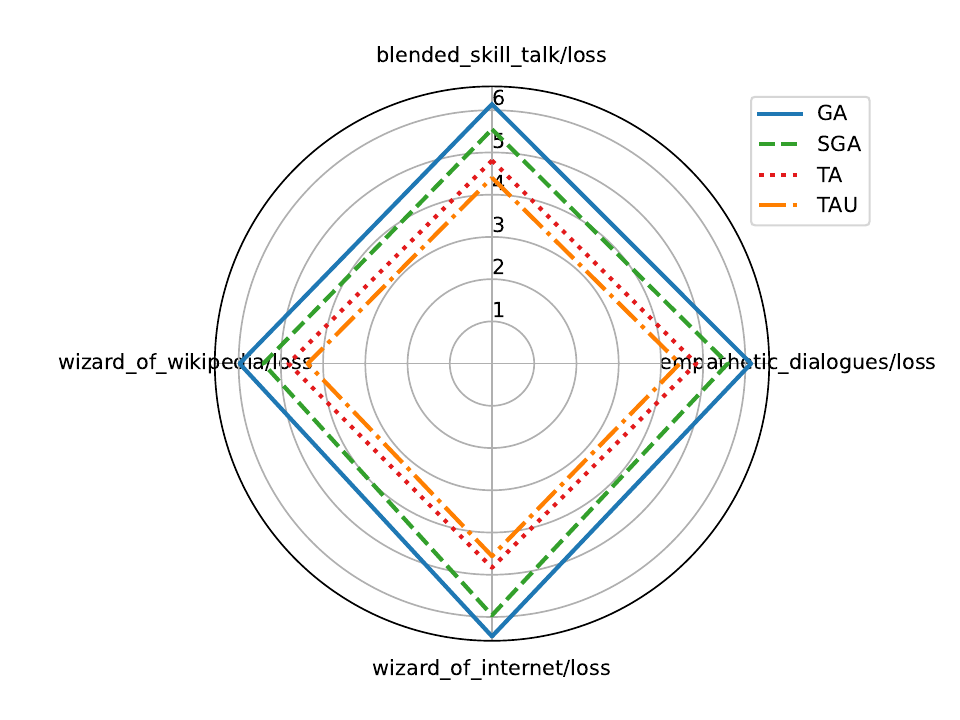}
  \includegraphics[width=0.33\linewidth]{ 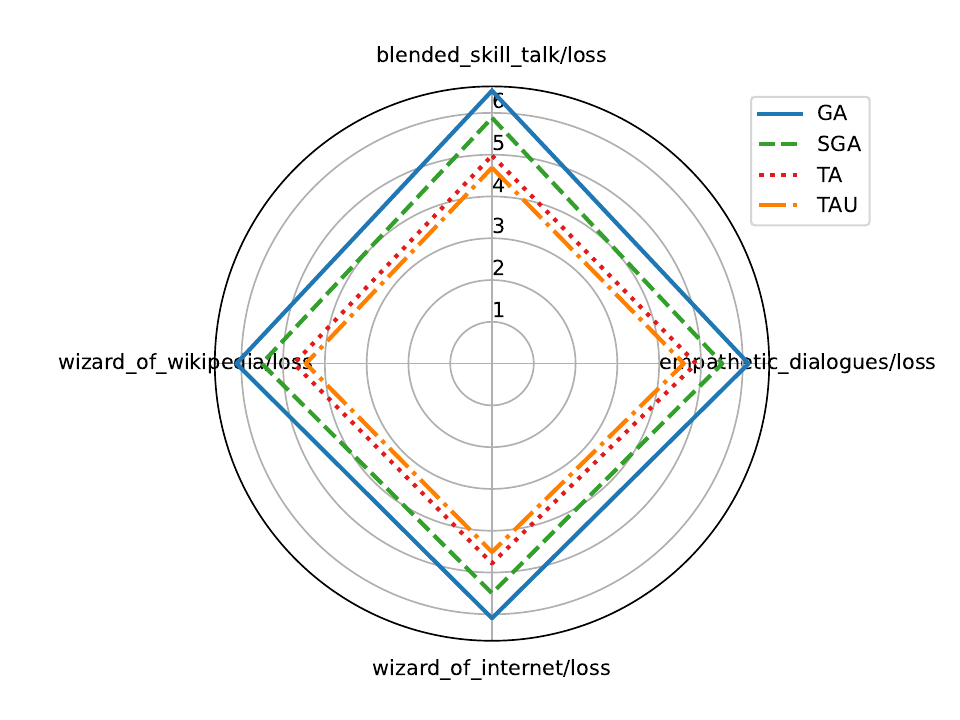}
}

\caption{The average loss across the four dialogue tasks. From left to right: unlearning 8 samples, 16 samples, and 32 samples at once, respectively. Results based on the 125 million checkpoint. }
\end{center}
\vskip -0.2in
\end{figure*}

\pagebreak

\begin{figure*}[ht]
\vskip 0.2in
\begin{center}
\centerline{
  \includegraphics[width=0.33\linewidth]{ 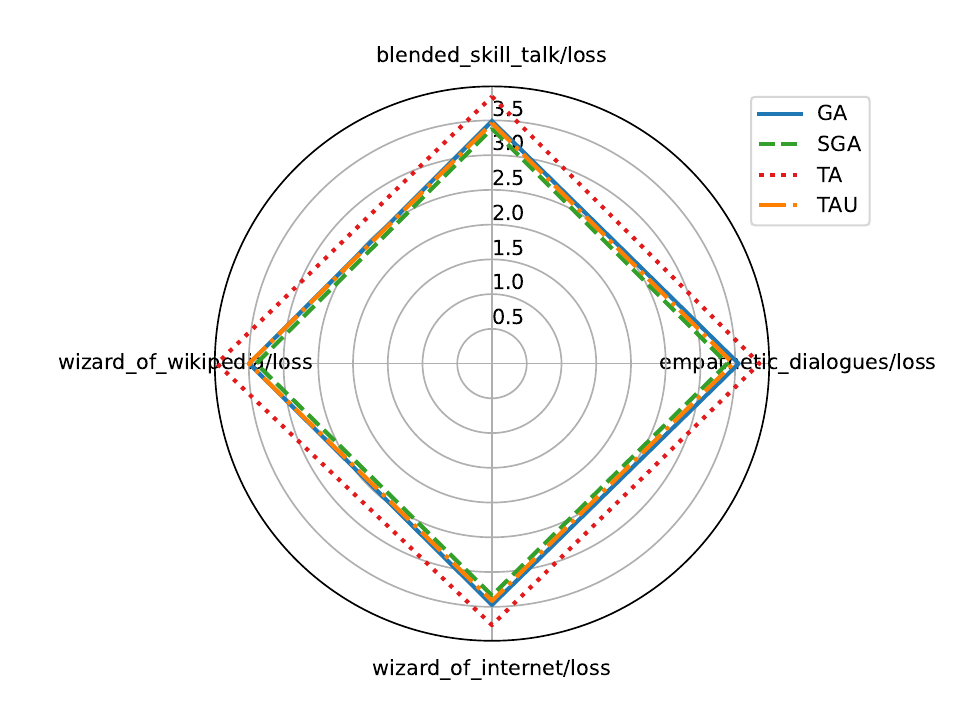}
  \includegraphics[width=0.33\linewidth]{ 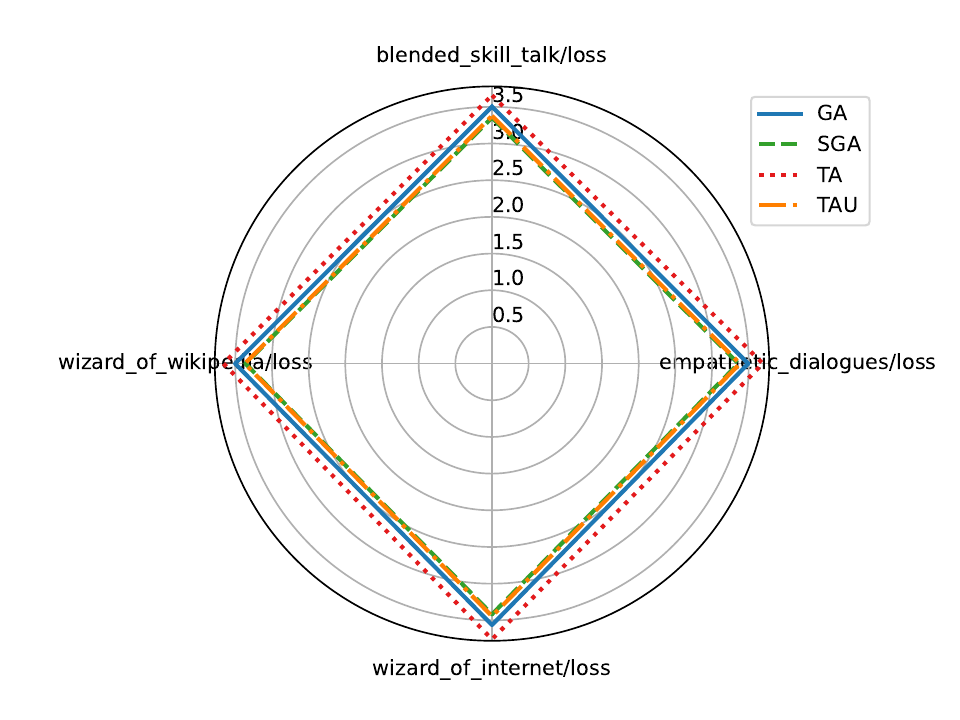}
  \includegraphics[width=0.33\linewidth]{ 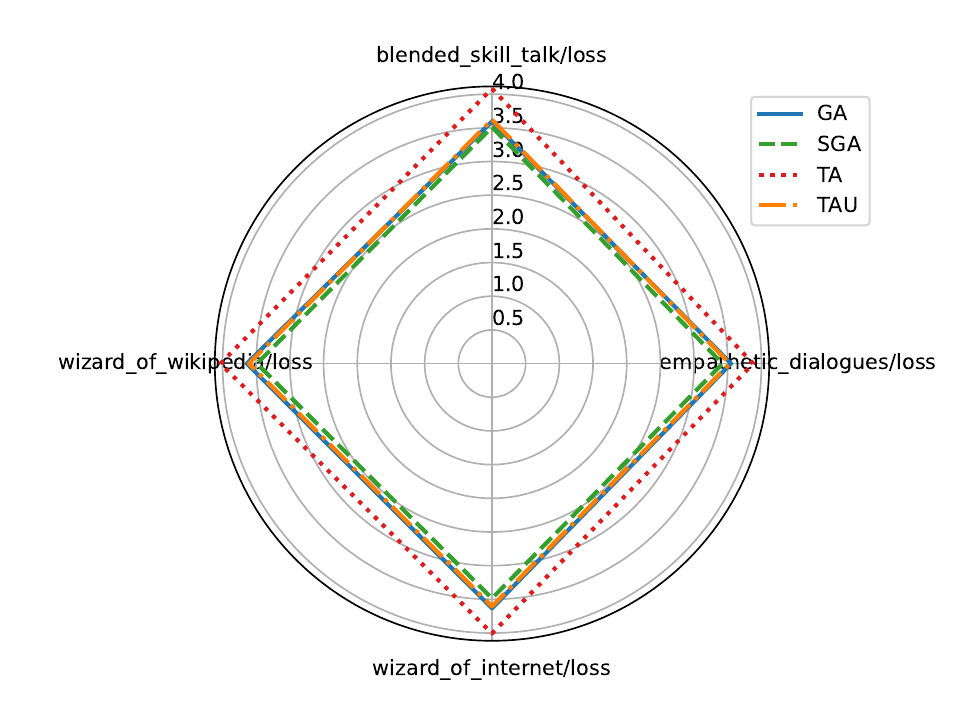}
}

\caption{The average loss across the four dialogue tasks. From left to right: unlearning 8 samples, 16 samples, and 32 samples at once, respectively. Results based on the 1.3B checkpoint. }
\end{center}
\vskip -0.2in
\end{figure*}

\begin{figure*}[ht]
\vskip 0.2in
\begin{center}
\centerline{
  \includegraphics[width=0.33\linewidth]{ 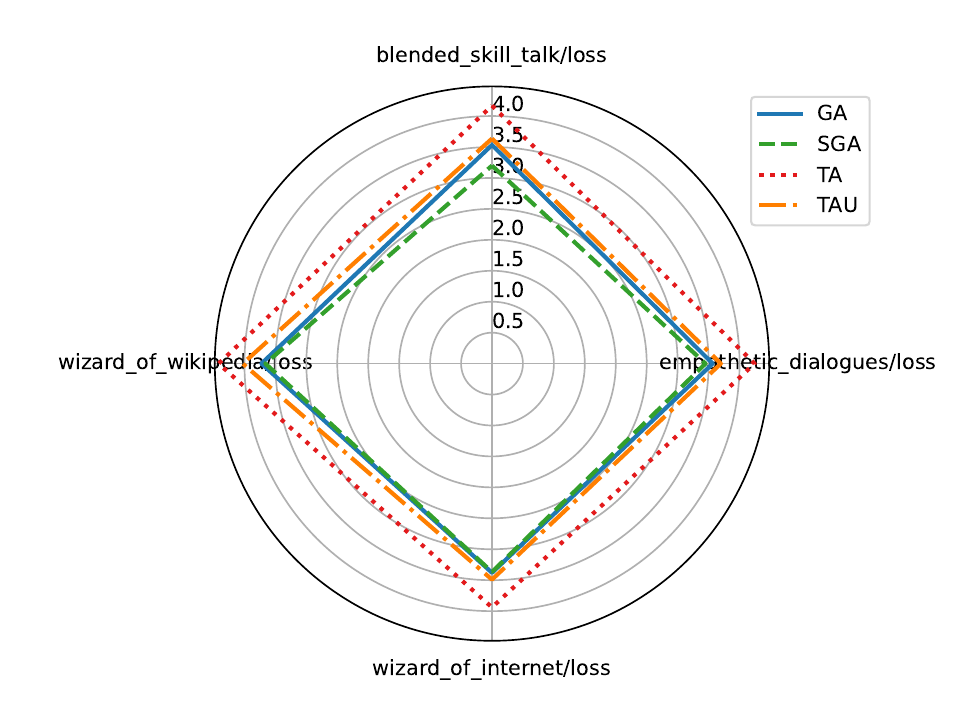}
  \includegraphics[width=0.33\linewidth]{ 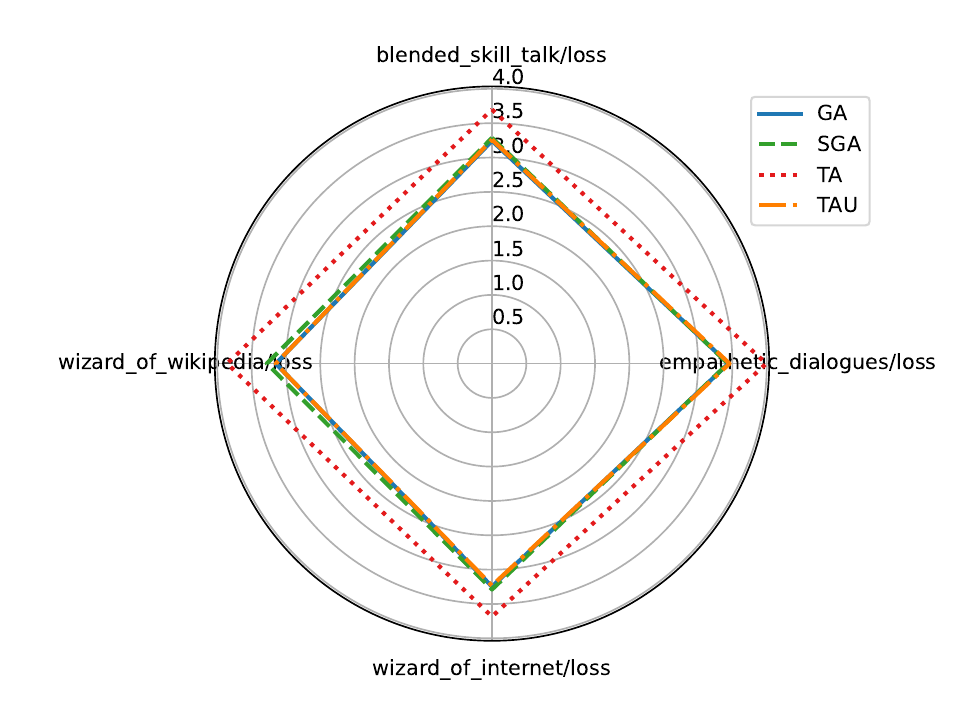}
  \includegraphics[width=0.33\linewidth]{ 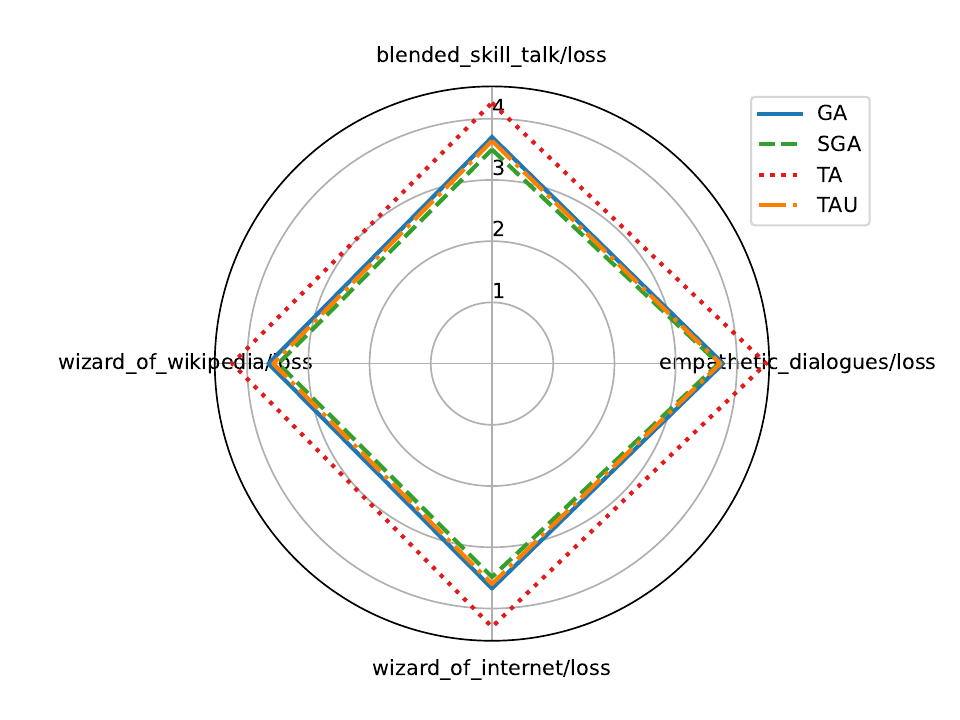}
}

\caption{The average loss across the four dialogue tasks. From left to right: unlearning 8 samples, 16 samples, and 32 samples at once, respectively. Results based on the 2.7B million checkpoint.}
\end{center}
\vskip -0.2in
\end{figure*}

\end{document}